\documentclass[dvipsnames]{article} %
\usepackage{colm2024_conference}

\usepackage{booktabs}

\usepackage{enumitem}
\usepackage{wrapfig}


\usepackage{algorithm}
\usepackage{algpseudocode}

\usepackage{graphicx}
\usepackage[misc]{ifsym}
\usepackage{mathtools}   

\usepackage{microtype}
\usepackage{colortbl}
\usepackage[utf8]{inputenc}
\usepackage[T1]{fontenc}
\definecolor{lightgray}{rgb}{0.9,0.9,0.9}
\usepackage{caption}
\usepackage{subcaption}
\usepackage{setspace}
\usepackage{url}
\usepackage{multirow}
\usepackage{tabularx}
\usepackage{blindtext}
\usepackage{pgfplots}
\pgfplotsset{compat=1.18} 
\usepackage{tikz}
\usetikzlibrary{er,positioning,bayesnet}
\usepackage{makecell}
\usepackage{tipa}
\usepackage{siunitx}
\usepackage{nicefrac}
\usepackage{listings}
\usepackage[raster,skins, most]{tcolorbox} %
\usepackage{xltabular}
\usepackage{adjustbox}
\usepackage{xurl}
\usepackage{rotating}
\usepackage[normalem]{ulem}
\usepackage{xspace}

\usepackage{CJKutf8}

\usepackage{fontawesome}
\usepackage{pifont}
\usepackage{fancyvrb}
\usepackage{color}
\definecolor{green_first}{RGB}{168, 209, 176}   
\definecolor{green_second}{RGB}{200, 235, 200}  
\definecolor{green_third}{RGB}{235, 255, 235}   
\definecolor{light_green_table}{RGB}{220, 255, 220}  
\definecolor{light_purple_table}{RGB}{235, 225, 255}  
\definecolor{light_green}{rgb}{0.569, 0.800, 0.459}
\definecolor{blue_dist}{rgb}{0.192,0.443,0.651}
\definecolor{dark_purple_table}{RGB}{103, 74, 241}  
\definecolor{purple_cite}{RGB}{1, 2, 129}  

\definecolor{purple_dist}{RGB}{125,94,237}

\definecolor{orange_dist}{rgb}{0.812,0.545,0.239}
\definecolor{yellow_dist}{rgb}{0.918,0.804,0.463}
\definecolor{website}{rgb}{0.9333333333333333, 0.10980392156862745, 0.592156862745098} 
\definecolor{pm_rowcolor}{rgb}{0.85, 0.90, 0.84} 
\definecolor{medium_gray}{RGB}{150, 150, 150}  
\definecolor{medium_purple}{RGB}{150, 120, 200}  
\newtcolorbox{promptbox}[2][Prompt]{
    breakable,
    colback=black!5!white,        
    arc=5pt,                      
    boxrule=0.5pt,                
    fonttitle=\bfseries,          
    title=#1,                     
    before upper={\small},        
    fontupper=\fontfamily{ptm}\selectfont, 
    colframe=#2,                  
    left=3pt,                     
    right=3pt,                    
    top=3pt,                      
    bottom=3pt,                   
    boxsep=3pt,                   
    toptitle=1pt,                 
    bottomtitle=1pt,              
    lefttitle=1pt,                
    righttitle=1pt,               
}

\usepackage{xcolor}      
\newcommand\inb[1]{\colorbox{gray!20}{\lstinline|#1|}}

\useunder{\uline}{\ul}{}

\usepackage{amsmath,amsfonts,bm}

\def\eqref#1{equation~\ref{#1}}

\def\1{\bm{1}}

\DeclareMathAlphabet{\mathsfit}{\encodingdefault}{\sfdefault}{m}{sl}
\SetMathAlphabet{\mathsfit}{bold}{\encodingdefault}{\sfdefault}{bx}{n}

\usetikzlibrary{tikzmark}
\makeatletter
\newcommand*\myfontsize{%
  \@setfontsize\myfontsize{7}{8}%
}
\makeatother

\definecolor{uclablue}{RGB}{159, 195, 224}

\definecolor{uclagold}{RGB}{255, 240, 180}

\definecolor{aliceblue}{RGB}{255, 238, 241}

\definecolor{cadmiumgreen}{rgb}{0.0, 0.42, 0.24}

\definecolor{myred}{rgb}{0.7, 0.3, 0.0}
\definecolor{myblue}{rgb}{0.2, 0.3, 0.6}
\definecolor{babygreen}{rgb}{0.85, 0.97, 0.85}

\definecolor{purple1}{RGB}{126, 107, 196}
\definecolor{purple2}{RGB}{199, 158, 207}
\definecolor{purple3}{RGB}{214, 200, 255}
\definecolor{purple4}{RGB}{254, 240, 255}

\definecolor{deepblue}{RGB}{48, 58, 82}

\definecolor{deepPurple}{HTML}{330066}

\definecolor{uclablue_old}{rgb}{0.15, 0.45, 0.68}
\hypersetup{
    breaklinks,
    citecolor=uclablue_old,
    colorlinks=true,
}

\newtcolorbox{mybox}[2][]
  {colback = black!5!white, colframe = black!75!black, fonttitle = \bfseries,
    colbacktitle = black!100!black, enhanced, before upper={\fontsize{8}{11}\obeyspaces\obeylines\selectfont}, fontupper=\selectfont,
    attach boxed title to top left={yshift=-2.2mm,xshift=4mm},
    title=#2,#1}

\title{%
\begin{tabular}[t]{l}
  \parbox[t]{1.0\textwidth}{\centering
    SocialArena: Evaluating Social Intelligence in Large Language Models
  }
\end{tabular}
}

\author{
Zongqi Wang$^{1,3}$\thanks{Work done during internship at Alibaba Token Hub, Alibaba Group.},
Rui Wang$^{1}$,
Yuchuan Wu$^{1}$,
Hua Yang$^{2}$,
Bingxue Qiu$^{2}$,
Tianle Gu$^{3}$, \\
Bing Zhao$^{2}$,
Yujiu Yang$^{ (\textrm{\Letter}),3}$,
Jieping Ye$^{2}$,
Yongbin Li$^{ (\textrm{\Letter}),1}$
  \\[1em]
  {\fontsize{10pt}{11pt}\selectfont $^{1}$Alibaba Token Hub, Alibaba Group, $^{2}$Alibaba Group, $^{3}$Tsinghua University}
}

\begin{document}

\maketitle

\begin{abstract}
We introduce SocialArena, a benchmark for systematically defining and evaluating the social intelligence of large language models (LLMs). Although LLMs are increasingly deployed in socially interactive applications such as mental health counseling, companionship chatbots, and educational tutoring, existing evaluations remain fragmented—focusing on isolated capabilities or single scenarios—and evaluation reliability is largely unverified. 
This work makes three main contributions to address these limitations. 
(1) We first propose a unified three-tier framework that defines social intelligence along three capability dimensions: role-playing ability, mental state reasoning, and behavioral reasoning. To operationalize this definition, we design a role-playing dialogue paradigm with explicit character thinking for evaluation, enabling simultaneous assessment of these capabilities in a manner that mirrors realistic human interactions. 
(2) To ensure diversity across social scenarios and thereby prevent evaluation conclusions from being confined to specific scenarios, we introduce the U-TRIMF framework, which guides us in synthesizing social contexts with broad coverage. 
(3) To obtain reliable evaluation, we invest \$132.4K to recruit and train human annotators and adopt a human-annotated arena-style protocol, where annotators compare model responses and provide preference labels along with natural-language critiques. 
%
We hope that SocialArena can provide valuable insights and guidance for future LLM improvements. 

\end{abstract}


\section{Introduction}

As large language models (LLMs) continue to grow in capability, they are increasingly entering the human world. 
Existing work has demonstrated that LLMs possess strong problem-solving abilities~\citep{shao2025deepseekmath,jimenez2023swe,barres2025tau}, such as mathematical reasoning and code generation. Yet the human world is not solely about problem-solving—much of it revolves around the capacity for interaction with humans, such as understanding human thoughts and emotions, navigating social relationships, and responding with emotional intelligence. This capacity, broadly referred to as \textit{social intelligence}, is indispensable for LLMs to enter the human world, with applications such as psychological counseling~\citep{zhang2025sentient,zhao2024esc}, education~\citep{markel2023gpteach}, and game NPCs~\citep{park2023generative}. However, despite its importance, how well current LLMs perform on social intelligence remains an open question. 
In this paper, we introduce \textbf{SocialArena}, a foundational benchmark that provides the first systematic and comprehensive definition and evaluation of social intelligence for LLMs. Specifically, we tackle the following critical challenges:



\definecolor{darkgreen}{rgb}{0.463, 0.726, 0}
\definecolor{lightgray}{rgb}{0.85, 0.85, 0.85}
\newcommand{\bad}{{\color{red}\ding{55}}}
\newcommand{\good}{{\color{darkgreen}\ding{51}}}
\newcommand{\na}{{\color{lightgray}\textbf{--}}}

\newcommand{\human}{\raisebox{-1.5pt}{\includegraphics[height=1em]{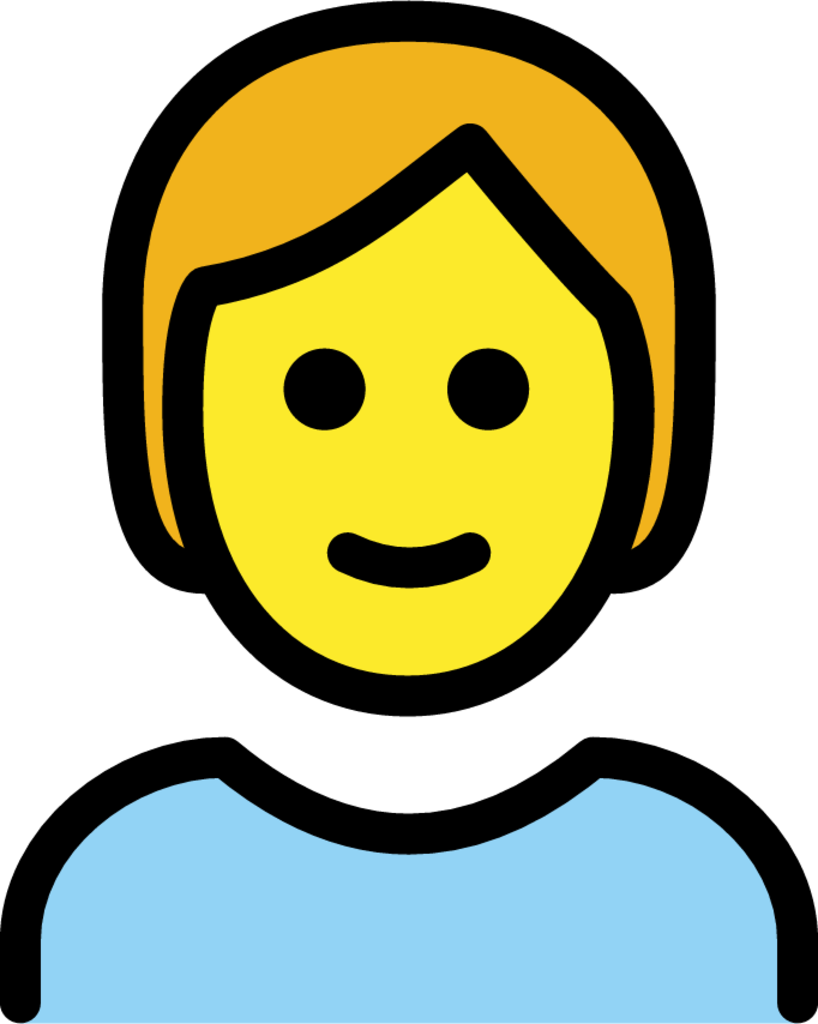}}}
\newcommand{\robot}{\raisebox{-1.5pt}{\includegraphics[height=1em]{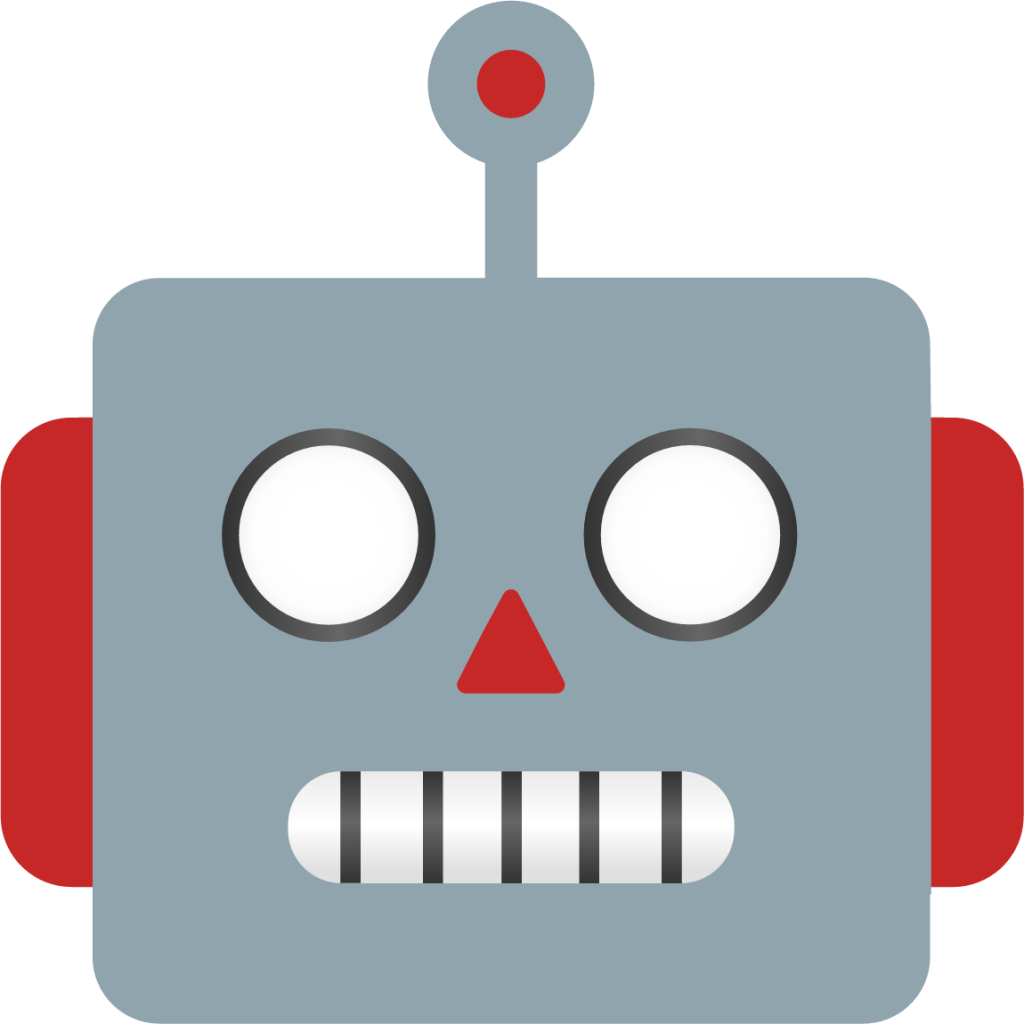}}}
\newcommand{\mc}{\raisebox{-1.5pt}{\includegraphics[height=1em]{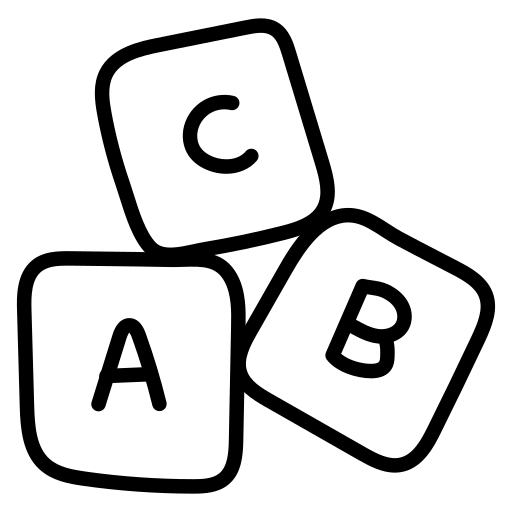}}}

\begin{table*}[t]
\centering
\caption{\textbf{Representative social intelligence benchmarks.} From left to right, the columns represent: whether it supports (\good) or lacks (\bad) the evaluation of role-playing capability, ToM reasoning for seven mental states and contextually appropriate behavior reasoning; the number of prototypes and samples of social scenarios; and annotators (\human: human annotation; \robot: LLM-as-a-Judge \mc: multiple-choice).}
\begin{adjustbox}{max width=\textwidth, scale=1}
\begin{tabular}{ll c ccccccc c cc c}
    \toprule
    \multirow{3}{*}{\textbf{Dataset}} & 
    \multirow{3}{*}{\textbf{Date}} & 
    \multirow{3}{*}{\textbf{Roleplay}} & 
    \multicolumn{7}{c}{\textbf{ToM Reasoning}} & 
    \multirow{3}{*}{\textbf{Behavior}} & 
    \multicolumn{2}{c}{\textbf{Social Context}} & 
    \multirow{3}{*}{\textbf{Annot.}} \\
    \cmidrule(lr){4-10} \cmidrule(lr){12-13}
    & & & \textbf{Emo.} & \textbf{Des.} & \textbf{Int.} & \textbf{Per.} & \textbf{Kno.} & \textbf{Bel.} & \textbf{Non-lit.} & & \textbf{\#Proto.} & \textbf{\#Samples} & \\
    \midrule
    \multicolumn{14}{l}{\textit{Role-playing Benchmarks}} \\
    InCharacter~\citep{wang2024incharacter} & Oct 23 & \good & \na & \na & \na & \na & \na & \na & \na & \na & \na & 1,408 & \robot \\
    CharacterBench~\citep{zhou2025characterbench} & Dec 24 & \good & \na & \na & \na & \na & \na & \na & \na & \na & \na & 3,250 & \robot \\
    CoSER~\citep{wang2025coser} & Feb 25 & \good & \good & \good & \good & \bad & \good & \good & \bad & \good & 1 & 200 & \robot \\
    RMTBench~\citep{xiang2025rmtbench} & Jul 25 & \good & \good & \bad & \bad & \bad & \bad & \bad & \bad & \good & 1 & 160 & \robot \\
    \midrule
    \multicolumn{14}{l}{\textit{Theory-of-Mind Benchmarks}} \\
    EmoBench~\citep{sabour2024emobench} & Feb 24 & \bad & \good & \bad & \bad & \bad & \bad & \bad & \bad & \good & \na & 400 & \mc \\
    OpenToM~\citep{zhou2025characterbench} & Feb 24 & \bad & \bad & \bad & \good & \bad & \bad & \good & \bad & \bad & \na & 13,708 & \mc \\
    SimpleToM~\citep{gu2024simpletom} & Oct 24 & \bad & \bad & \bad & \bad & \good & \bad & \bad & \bad & \bad & \na & 3,441 & \mc \\
    ToMATO~\citep{shinoda2025tomato} & Jan 25 & \bad & \good & \good & \good & \bad & \good & \good & \bad & \bad & \na & 5,400 & \mc \\
    \midrule
    \multicolumn{14}{l}{\textit{Social Intelligence Benchmarks}} \\
    SocialIQA~\citep{sap2019socialiqa} & Sep 19 & \bad & \na & \na & \na & \na & \na & \na & \na & \bad & \na & 2,000 & \mc \\
    SOTOPIA~\citep{zhou2023sotopia} & Oct 23 & \good & \bad & \bad & \good & \bad & \bad & \bad & \bad & \good & 6 & 450 & \robot \\
    SocialBench~\citep{chen2024socialbench} & Feb 24 & \good & \na & \na & \na & \na & \na & \na & \na & \bad & \na & 6,000 & \mc \\
    EQ-Bench3~\citep{eqbench3_repo_2025} & Jan 25 & \good & \good & \good & \bad & \bad & \bad & \good & \bad & \bad & 3 & 45 & \robot \\
    SAGE~\citep{zhang2025sentient} & Feb 25 & \good & \good & \bad & \good & \bad & \bad & \bad & \bad & \good & 1 & 100 & \robot \\
    SI-Bench~\citep{huang2025si} & Mar 25 & \good & \good & \good & \bad & \bad & \good & \bad & \bad & \good & 1 & 312 & \human \\
    \midrule
    \textbf{SocialArena (ours)} & May 26 & \good & \good & \good & \good & \good & \good & \good & \good & \good & 600 & 5,702 & \human \\
    \bottomrule
\end{tabular}
\end{adjustbox}
\label{tab:comparison_of_social_benchmarks}
\end{table*}

\noindent \textbf{Q1: What is the definition of social intelligence, or what capabilities does it encompass?}

Currently, several efforts have attempted to evaluate social intelligence in LLMs; however, these evaluations remain fragmented, with each typically focusing on only one or a few isolated capability dimensions. 
To establish a clear definition of social intelligence, we provide a comprehensive review in Sec.~\ref{sec:related_work} (summarized in Tab.~\ref{tab:comparison_of_social_benchmarks}). Building upon it, we propose a unified three-tier capability framework that systematically defines social intelligence. Specifically, we decompose the capabilities underlying social intelligence into three dimensions:
(1) \textbf{Role-playing}: the capacity to maintain consistent linguistic style, emotional tone, and other character attributes~\citep{wang2024incharacter,zhou2025characterbench,lu2025rolemrc,wang2025coser}. This capability forms the foundation of a character. While humans exhibit such consistency naturally, LLMs may suffer from character inconsistency, making it necessary for evaluation;
(2) \textbf{Mental state reasoning}: the capacity to infer one's own and others' mental states, corresponding to Theory of Mind (ToM). This capability is a prerequisite for meaningful social engagement~\citep{xu2024opentom,gu2024simpletom,shinoda2025tomato,wu2023hi};
%
and (3) \textbf{Behavioral reasoning}: the capacity to reason about what response or action is most appropriate and how to execute it effectively~\citep{zhou2023sotopia,sabour2024emobench,zhang2025sentient,huang2025si}. Unlike role consistency and mental state reasoning, which is scenario-agnostic, behavior reasoning is largely scenario-dependent. 
%
\textit{In summary, we define social intelligence in LLMs as the ability to infer one's own and others' mental states, generate contextually appropriate behaviors based on the given social context, while maintaining role consistency throughout the interaction.}


%
\noindent \textbf{Q2: How can we ensure the evaluation covers a sufficiently broad range of social scenarios?}

As discussed above, different social scenarios require the model to produce different behaviors. Existing benchmarks, however, are usually designed to evaluate single scenario such as dating applications~\citep{huang2025si}, psychological counseling~\citep{zhang2025sentient}, business social interactions~\citep{zhou2023sotopia}, or virtual world narrative progression~\citep{wang2025coser,xu2026adamarp}. Consequently, the generalizability of evaluation conclusion of these benchmarks remains limited. 

To achieve broad scenario coverage, we propose a scenario synthesis pipeline centered on the \textbf{U-TRIMF} framework. 
U-TRIMF follows a top-down design: it first identifies six fundamental dimensions that jointly constitute a social scenario, and then enumerates exhaustive sub-categories along each dimension, thereby guaranteeing wide coverage of the scenario space (see Tab.~\ref{tab:trimf-main} for details of U-TRIMF). 
%
%
We refer readers to Sec.~\ref{sec:dataset_curation_pipeline} for a detailed description of how U-TRIMF is used.

\noindent \textbf{Q3: How do we ensure reliable evaluation?}

Existing evaluations of social intelligence mainly fall into two paradigms, both with notable limitations. Multiple-choice QA~\citep{sap2019socialiqa,sabour2024emobench,gu2024simpletom} reduces situated social interactions to static, decontextualized selections, failing to capture the generative nature of real-world behavior. 
LLM-as-a-Judge~\citep{zheng2023judging} offers scalability, but as shown in Sec.~\ref{sec:eval_llm_as_a_judge}, current LLMs still misalign substantially with human judgments. Given these limitations, we adopt a human-centered protocol: we recruit and train human annotators and design an arena-style comparison framework with explicit critique collection, ensuring both scalable rankings and trustworthy, diagnostically meaningful assessments.

\begin{figure*}[!t]
\centering
    \includegraphics[width=1.0\textwidth]{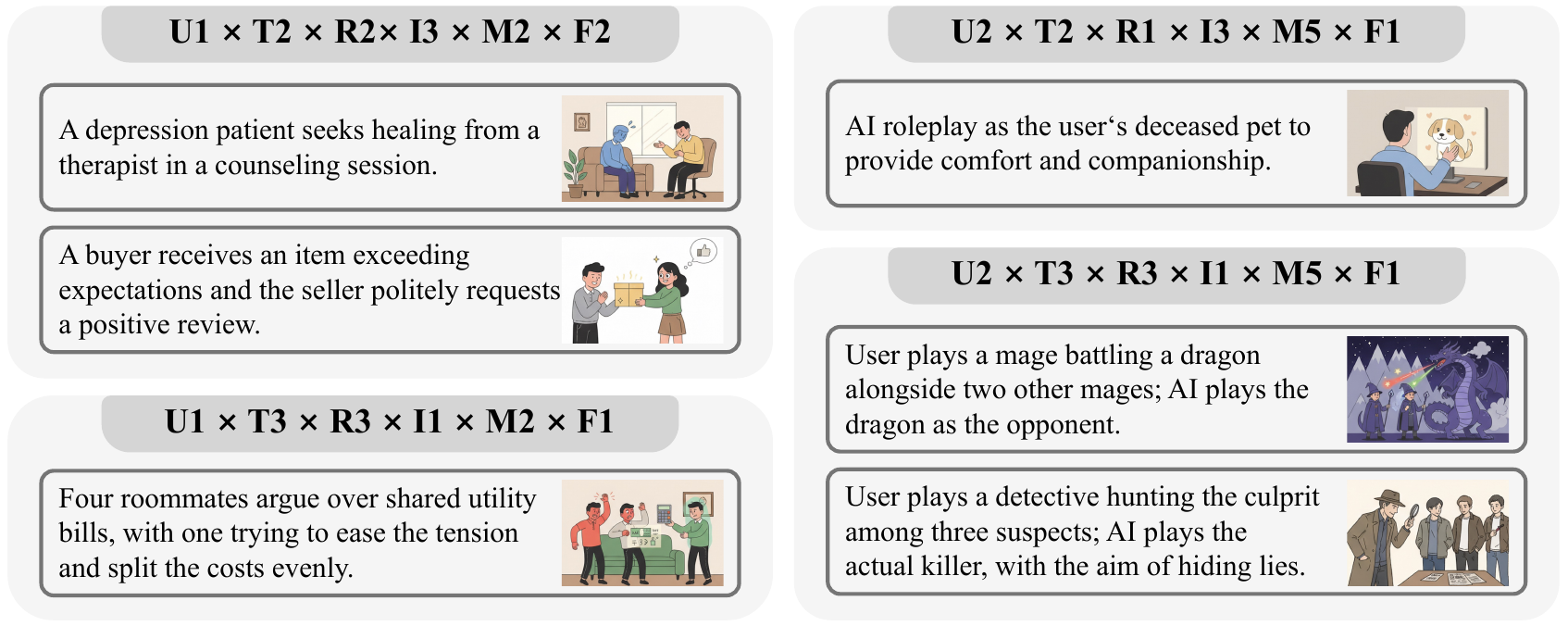}
\caption{Examples of base scenario templates constructed by \textbf{\textsc{U-TRIMF}} framework.}
\label{fig:examples_base_templates}
\end{figure*}

\noindent \textbf{Q4: How do current mainstream LLMs perform on social intelligence?}


Our experiments reveal the complexity of LLMs' social intelligence and suggest several future directions. 
First, all LLMs show a consistent bias: their behavioral reasoning is much weaker than social understanding. This gap highlights their fragility in turning social perception into action and decisions. 
Second, models exhibit distinct strengths across capability dimensions, with no model consistently dominating all aspects. 
Third, model rankings vary across social contexts, indicating limited generalization across scenarios. 
We hope SocialArena enables fine-grained diagnosis of socio-cognitive deficiencies and drives future advances in LLM social intelligence.

\section{SocialArena}

\subsection{Benchmark Overview}

SocialArena is designed as a comprehensive benchmark to evaluate social intelligence in LLMs across three capabilities: \textbf{Role-playing}, \textbf{Mental State Reasoning}, and \textbf{Behavioral Reasoning}. 
SocialArena follows an arena-style paradigm~\citep{zheng2023judging}, where anonymous models are paired against each other within the same social context, and their responses are ranked via human voting, with leaderboard based the Elo~\citep{elo1966uscf} rating. 
SocialArena is a human-based benchmark that relies on human votes. We recruit human annotators to label the dataset to establish the leaderboard. 
In total, this dataset comprises \textbf{5,702 social scenarios} and \textbf{17,676 human votes}. 

\begin{table*}[t]
\centering
\small
\caption{\textbf{\textsc{U-TRIMF} framework for base scenario construction.} Each scenario is decomposed into three core elements: \textbf{(1) User Presence} — whether a user exists in the scenario, which fundamentally shapes the LLM's role and responsibilities: without a user, the LLM acts as a character embedded in and governed by the real world; with a user, the LLM need to serve user-specific needs, and gain the authority to control world evolution; \textbf{(2) Characters} — how many characters are involved, their relationships, and interest structures; \textbf{(3) Events} — what happens and under what situational context. See Tab.~\ref{tab:trimf} for details.}
\begin{tabular}{m{3.0cm}m{7.5cm}m{4.0cm}}
\toprule
\textbf{Dimension} & \textbf{Sub-categories} & \textbf{Source} \\
\midrule
\textbf{U}ser Presence & U1: Real-world \quad U2: User-centered & --- \\
\midrule
\textbf{T}opological Relation & \sout{T1: Individual} \quad T2: Dyad \quad T3: Group & Topology~\citep{kluver1999topology} \\
\midrule
\textbf{R}elational Model & R1: Communal Sharing \quad R2: Authority Ranking \quad R3: Equality Matching \quad R4: Market Pricing \quad R5: Null & Relational Models Theory~\citep{fiske1992four} \\
\midrule
\textbf{I}nterest Structure & I1: Zero-sum \quad I2: Negative-sum \quad I3: Positive-sum & Game Theory~\citep{fudenberg1991game} \\
\midrule
\textbf{M}otivational Type & M1: Physiological \quad M2: Safety \quad M3: Belonging \quad M4: Esteem \quad M5: Self-actualization & Maslow's Hierarchy of Needs~\citep{mcleod2007maslow} \\
\midrule
\textbf{F}ormality & F1: Informal \quad F2: Formal & Social Formality~\citep{hollinger1982formal} \\
\bottomrule
\end{tabular}
\label{tab:trimf-main}
\end{table*}

\subsection{Social Scenarios Curation Pipeline}
\label{sec:dataset_curation_pipeline}

Social scenarios in SocialArena are synthesized following two important criteria: (1) ensuring comprehensive coverage and diversity of social contexts, and (2) ensuring the quality of scenarios. The model used for data synthesis is Qwen3-235B-A22B-Instruct-2507~\citep{qwen3technicalreport}. 

\textbf{Step 1: Base Scenario Template Construction.}
To ensure comprehensive coverage, we adopt a top-down strategy to construct base scenario templates using the \textsc{U-TRIMF} framework as shown in Tab.~\ref{tab:trimf-main}. The key insight is that by decomposing social scenarios into a set of orthogonal dimensions and systematically sampling from their sub-dimensions, we can achieve broad and structured coverage of the social space. 
When constructing a social scenario, we first uniformly sample one sub-category from each of the six dimensions and compose them into a base scenario template. We present some examples in Fig.~\ref{fig:examples_base_templates}. 

\textbf{Step 2: Keyword-Guided Scenarios Synthesis.}
Once the basic templates are established through the U-TRIMF framework, we proceed to generate concrete data instances. Each generated sample consists of \textit{a plot summary, current scene description with optional character descriptions, dialogue history, and the next speaker}. To further enhance diversity, we incorporate 10 keywords randomly sampled from a large corpus (refer to \S~\ref{appendix:keyword_guided}) as additional soft guidance during generation. 

\textbf{Step 3: Quality Filtering.}
We iteratively develop several empirical rules (refer to \S~\ref{appendix:quality_filtering}) covering complexity, logical coherence, safety, and alignment with human values for preliminary automated filtering. Scenarios that pass this initial filter then undergo manual review to ensure final quality.

\subsection{Model Output}
\label{sec:model_output}

Given a social scenario, the next question is how to design the output format of the evaluated LLM.

\noindent We frame the model's output as a \textbf{role-playing dialogue with explicit character thinking}. 
This design not only aligns the evaluation process with real human social interactions, but also enables the simultaneous assessment of all three capability dimensions defined above. 
Specifically, the evaluated LLMs are required to role-play as \textit{a character in a given social context} and generate two components: an explicit \textit{character thinking} trace that reasons about the mental states of involved characters and analyzes the current scene to devise appropriate behavior, followed by the final output \textit{behavior (responses and actions)}. 

In a pilot annotation study, we found that free-form character thinking yields poor annotation quality. For mental state reasoning, unstructured traces often interleave inferences about different mental states in an arbitrary order and at arbitrary positions. For behavioral reasoning, the model may draw on any available information, follow arbitrary reasoning trajectories, and even backtrack or self-revise before arriving at a decision. Such disorganized traces make it hard for annotators to localize the relevant segments for each judgment dimension, ultimately degrading annotation quality. We therefore introduce two design choices to improve annotation quality. 

\textbf{(1) Structured mental state reasoning.} We require the model to organize its mental-state inferences along the seven ToM categories — \textit{emotions, desires, intentions, percepts, knowledge, beliefs,} and \textit{non-literal communication}. Since different scenarios typically involve different subsets of mental states, the model is not required to produce inferences for all seven categories. This taxonomy is both \textit{theoretically complete} and \textit{structured}, allowing annotators to evaluate each category individually and substantially reducing annotation difficulty. See more details about ToM in \S~\ref{appendix:tom_by_mental_states}. 

\textbf{(2) Outcome-only behavior annotation.} For behavioral reasoning, annotators are instructed to judge only the final behavior, regardless of how many reasoning steps or what reasoning strategies led to it. This evaluation strategy also aligns with the widely adopted outcome-reward paradigm in Reinforcement Learning from Verifiable Rewards (RLVR)~\cite{guo2025deepseek}. 

\begin{tcolorbox}[colback=white, colframe=black, title=An Example of Scenario in SocialArena]
\small

\textbf{[Plot Overview]}

Two former colleagues reunite, and the gap in their career paths puts their friendship under a subtle psychological strain.

\textbf{[Current Scene]}

Alex and Jamie used to be colleagues at the same tech company and got along well. Three years ago, Alex left to start his own business, while Jamie stayed.

Now Alex's company is doing well, having just completed a Series B funding round. Jamie has risen to a middle management position at the original company but is now facing departmental restructuring.

Today, they meet for a casual catch-up at a quiet café in the city center. Alex arrives first, orders coffee, and sits down. Jamie shows up a few minutes late.

Both value this friendship, but Alex's success stirs complicated feelings in Jamie—not jealousy, but a quiet re-examination of his own situation. Alex is also aware of this subtle tension and isn't sure whether to talk about work or avoid it.

\textbf{[Opening Dialogue]}

1. Jamie: (puts down bag, sits down) Have you been waiting long? Traffic was bad.

2. Alex: Just got here, coffee hasn’t come yet. How have you been?

3. Jamie: (shrugs) Same old, the company’s reorganizing again. What about you? Saw the news—your funding round went pretty well.

4. Alex: Not bad, just super busy. Honestly, funding isn’t all that glamorous—just more pressure.

5. Jamie: (smiles) Come on, getting funding is still better than being at the mercy of the bosses. Last month three people left our department, and they’re not replacing them.

6. Alex: Then how do you guys get the work done?

7. Jamie: Just tough it out. I’ve been working weekends for three weeks straight. The other day I was replying to emails at 2 a.m.

\textbf{[Character Information]}: Alex, Jamie

\textbf{[Next Speaker]}: Alex
\end{tcolorbox}

\begin{figure*}[h]
\centering
    \includegraphics[width=1.0\textwidth]{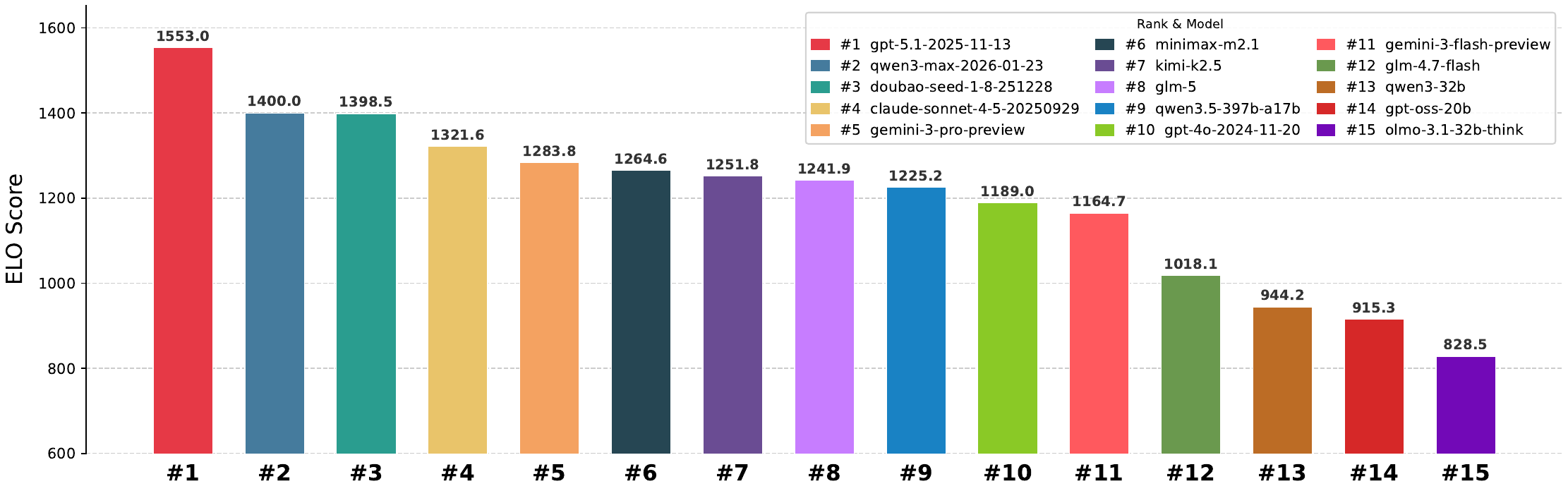}
\caption{Leaderboard of 15 frontier LLMs on SocialArena. Elo score is adopted. }
\label{fig:model_elo_ranking}
\end{figure*}

\subsection{Evaluation Protocol}
\label{sec:evaluation_protocol}

\paragraph{Task Formulation.}
Given a social context $S(p, b, h, n)$, where $p$ is the plot summary, $b$ is current scene description, $h$ is the dialogue history, and $n$ is the next speaker, two anonymized LLMs each generate a single-turn response $y_A$ and $y_B$ as character $n$. 
Annotators or platform users then evaluate each pair, providing a preference label $l \in \{A, B, T, G\}$, where $A$/$B$ indicates the superior response, $T$ denotes both are good, and $G$ denotes both are poor, along with optional natural-language critiques as justification. The leaderboard is computed from Elo~\citep{elo1966uscf} ratings derived from these pairwise outcomes. 

In our annotation process, we employ two strategies to enhance the quality and value of the annotations. 
First, for quality assurance, \textbf{each sample is independently annotated by at least three annotators}, with final preferences determined by majority voting; samples where all three annotators disagree are discarded. 
Second, to enable deeper interpretable analysis, \textbf{annotators are strictly required to provide natural-language critiques}, and to provide separate annotations across nine dimensions — the seven ToM mental state dimensions ($T_1$--$T_7$), overall character thinking quality ($T$), and behavior appropriateness ($B$) — rather than a single holistic judgment. 
Details on annotator recruitment, annotation guidelines, and quality control are provided in \S~\ref{appendix:evaluation_protocol}.

\section{Experiments}

\subsection{Evaluation Setup}

We select 15 models and conduct pairwise evaluations between every two models, with each pair engaging in 50 to 80 distinct battles. Each battle is evaluated by at least three independent annotators, and the final verdict is determined by majority voting. All models are then subsequently ranked using the Elo rating system~\cite{elo1966uscf}. 

\begin{figure*}[t]
\centering
    \includegraphics[width=1.0\textwidth]{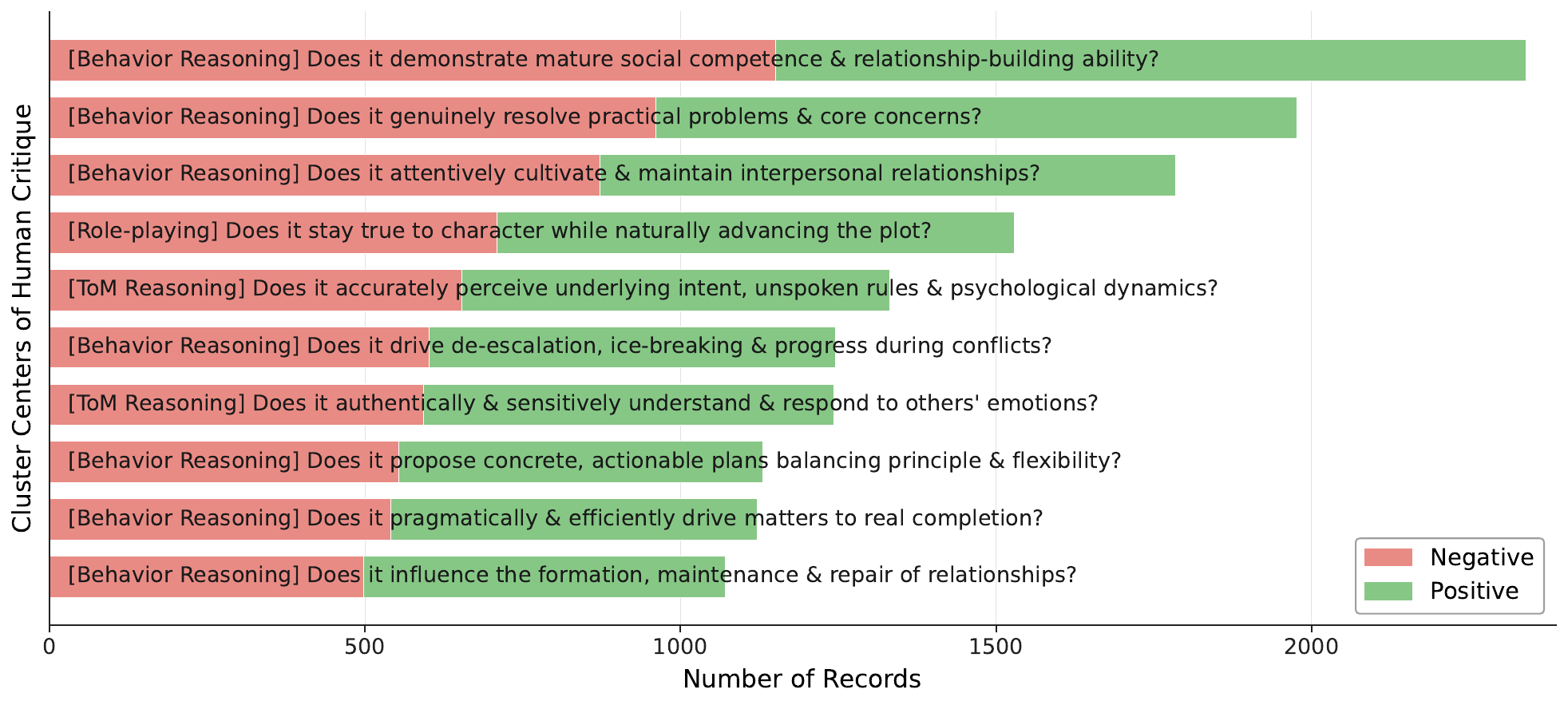}
\caption{Top 10 cluster centers of human critiques.}
\label{fig:evaluation_criteria_chart}
\end{figure*}

\subsection{Leaderboard}

Fig.~\ref{fig:model_elo_ranking} presents the Elo-based leaderboard. 
The top three positions are occupied by closed-source models—\texttt{gpt-5.1-2025-11-13}, \texttt{qwen3\--max-2026-01-23}, and \texttt{doubao\--seed\--1\--8\--251228} with comparable scores. 
Open-source models follow in the middle ranks, while smaller-scale models such as \texttt{qwen3-32b}, \texttt{gpt-oss-20b}, and \texttt{olmo-3.1-32b-think} rank at the bottom. 
An intriguing observation is that, although some models perform comparably to others on STEM, coding, and agentic benchmarks (e.g., \texttt{qwen3.5-397b-a17b} vs. \texttt{qwen3-max-2026-01-23}~\cite{qwen35blog}, \texttt{glm-5} vs. \texttt{claude-sonnet-4-5-20250929}~\cite{zeng2026glm}), they lag substantially on SocialArena. This suggests that social intelligence is largely orthogonal to problem-solving abilities and does not automatically emerge from training focused on primarily or exclusively on cognitive tasks.

\subsection{A Deeper Understanding of LLM Performance in Social Intelligence}
\label{sec:deeper_understanding}

\begin{figure*}[t]
\centering
    \includegraphics[width=1.0\textwidth]{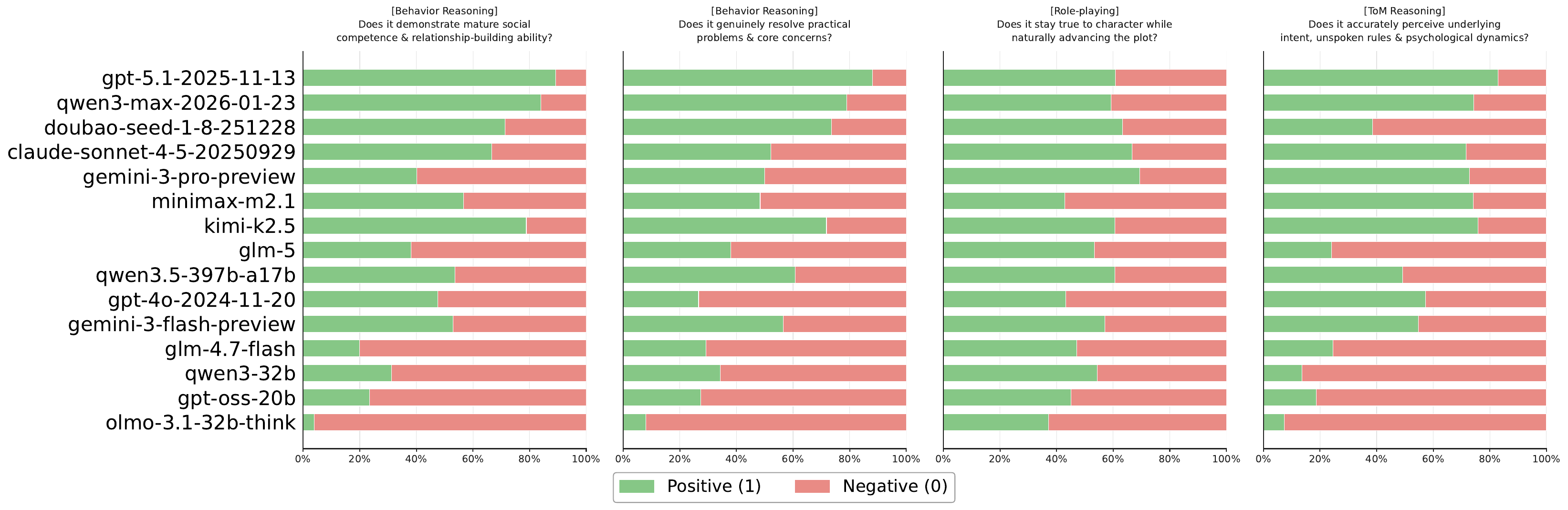}
\caption{Proportions of positive and negative human critiques for different models.}
\label{fig:model_comparison_chart_combined}
\end{figure*}

Beyond aggregate ranking scores, our dataset provides rich human critiques that enable a deeper diagnosis of model deficiencies. To support this analysis, we construct a structured taxonomy of evaluation rubrics derived directly from the collected human critiques, ultimately yielding \textbf{100 representative evaluation rubrics}. The detailed annotation procedure, rubric taxonomy construction pipeline, and full experimental results are presented in \S~\ref{appendix:deeper_understanding}. 

\paragraph{Frequently Occurring Dimensions Reveal Systematic Model Failures.} The frequency with which a dimension appears in human critiques serves as a dual signal: it reflects both the aspects of social intelligence that human annotators attend to most closely and the dimensions on which models most consistently fall short. 
As shown in Fig.~\ref{fig:evaluation_criteria_chart}, we draw the following conclusions.
\textit{\textbf{(1) Behavior Reasoning attracts the most attention, revealing that the true bottleneck of social intelligence lies in acting, not understanding.}} Six out of the top 10 cluster centers fall under Behavior Reasoning, substantially outnumbering ToM-related dimensions. This distribution exposes a counter-intuitive pattern: current models' deficiency is not that they fail to perceive social cues, but that they fail to translate such perception into actions that resolve problems, repair relationships, or drive matters to completion. Social intelligence evaluation should therefore shift its focus from the perception layer to the decision-and-action layer.
\textit{\textbf{(2) Negative critiques systematically dominate positive ones across nearly all high-frequency dimensions, indicating structural rather than sporadic failures.}} The asymmetry is especially pronounced on long-horizon, goal-oriented dimensions such as driving matters to real completion and de-escalating conflicts. 
%
\textit{\textbf{(3) Human evaluators judge social intelligence by relational outcomes, not by response correctness, exposing a paradigm-level mismatch with existing NLP evaluation.}} Dimensions such as cultivating and maintaining relationships and influencing relationship formation, maintenance, and repair recur among the top clusters. Unlike conventional criteria (helpfulness, harmlessness, factuality) that assume single-turn correctness, human judgments hinge on whether a relationship is advanced or repaired over time. This mismatch motivates a fundamental rethinking of how social intelligence should be benchmarked.
\textit{\textbf{(4) Balanced performance on Role-playing, contrasted with skewed negativity on reasoning dimensions, reveals a capability illusion.}} Role-playing—staying in character and advancing the plot—shows a relatively balanced positive/negative distribution, whereas Behavior and ToM Reasoning are overwhelmingly negative. Models thus appear socially competent on stylistic, surface-level consistency, but degrade sharply once evaluation probes deeper socio-cognitive reasoning. This implies that existing role-play-based social benchmarks may systematically overestimate models' true social intelligence.

\vspace{-1.2mm}
\paragraph{Divergent Per-Dimension Profiles Expose Model-Specific Weaknesses.} For each of the 100 dimensions, we compute the proportion of positive versus negative assessments received by each model, yielding a fine-grained per-dimension performance profile. As illustrated in Fig.~\ref{fig:model_comparison_chart_combined}, the performance profile of a single model fluctuates drastically across different evaluation dimensions, revealing sharp internal disparities between its strengths and weaknesses. Furthermore, when comparing across different models within the exact same dimension, the proportions of positive human critiques exhibit substantial variance, highlighting severe gaps in capability between top-tier and lower-performing architectures. 
\textit{\textbf{This observation leads to two important implications: (1) some models perform well on certain dimensions while exhibiting significant deficiencies on others, potentially stemming from insufficient or unbalanced training; and (2) evaluating social intelligence using only a small subset of pre-defined evaluation dimensions is inadequate. SocialArena offers a strong practical alternative by free-form natural-language human critiques.}}

\subsection{A Closer Look at Scenario-level Performance}

As discussed earlier, different social scenarios demand distinct model behaviors, implying that a single global ranking may not adequately reflect model suitability across all contexts. 
To investigate this, we compute pairwise Kendall $\tau$ rank correlations across typical scenarios, using randomly split halves of each scenario's data to estimate within-scenario ranking stability. 


\begin{figure}[t]
\centering
    \includegraphics[width=0.48\textwidth]{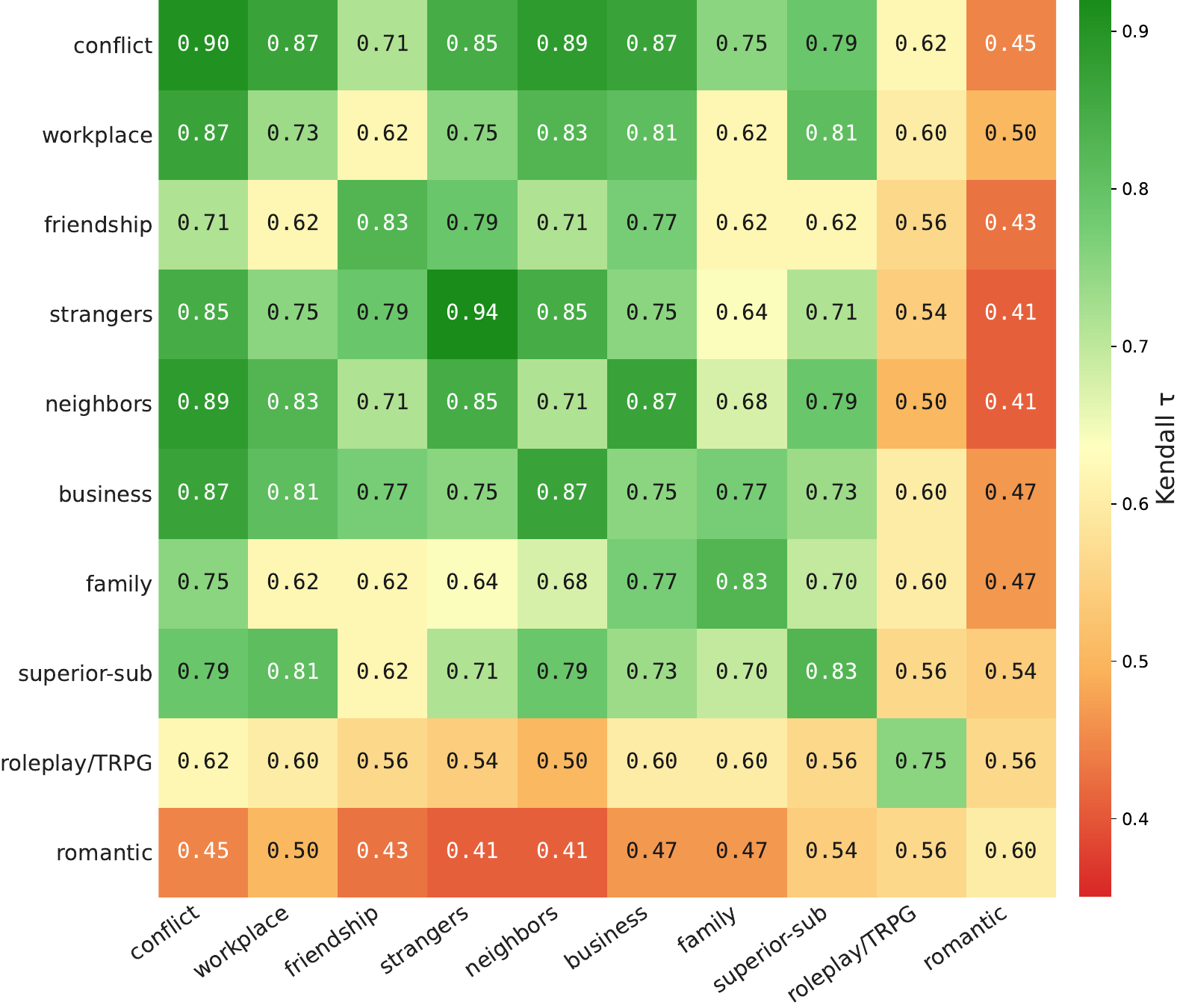}
\caption{Correlation heatmap of model rankings across typical scenarios.}
\label{fig:correlation_heatmap_rg}
\end{figure}

The results reveal two key findings. First, scenarios such as \textit{conflict}, \textit{friendship}, \textit{strangers}, \textit{family}, \textit{superior-subordinate}, \textit{roleplay/TRPG}, and \textit{romantic} exhibit high within-scenario rank consistency but low cross-scenario correlation, confirming that model rankings are indeed scenario-specific. This validates both the necessity of diverse scenario coverage for a comprehensive overall ranking and the importance of consulting scene-specific rankings when selecting models for targeted deployments. 

Second, not all scenarios show this pattern. \textit{Workplace}, \textit{neighbors}, and \textit{business} display rankings closely correlated with \textit{conflict}, rather than being self-contained. This is attributable to their high co-occurrence with conflict-tagged interactions, leading to substantial scenario overlap. Conversely, scenarios with low conflict frequency—such as \textit{romantic}, \textit{roleplay/TRPG}, and \textit{friendship}—show notably weaker correlation with \textit{conflict} rankings, further corroborating this explanation.

\section{Evaluation on LLM-as-a-Judge}
\label{sec:eval_llm_as_a_judge}

To investigate the reliability of LLM-as-a-Judge for such tasks, we construct \textbf{SocialArena-RewardBench}, a benchmark specifically designed to assess the alignment between LLM judges and human expert annotations in social intelligence evaluation. We randomly sampled a subset of instances from SocialArena. The samples are subsequently annotated and cross-validated by three human experts following the same annotation protocol described in Sec.~\ref{sec:evaluation_protocol}, resulting in a curated dataset of 152 samples. 

\paragraph{Evaluation Metric.}
Following~\citet{wang2026reward}, we evaluate annotation reliability along two complementary axes. \textbf{ACC@Label} measures whether the evaluator's preference label (A, B, T, or G) matches the human expert gold standard---capturing outcome-level agreement. \textbf{ACC@Critique} measures whether the evaluator's free-text critique is semantically consistent with the human reference critique, determined by whether the F1-based similarity between the predicted and gold-standard critique meets or exceeds a threshold of 0.5, following the same threshold setting as \citet{wang2026reward}. 

\begin{table}[t]
\centering
\small
\caption{Label accuracy and critique consistency of LLM-as-a-Judge on SocialArena-RewardBench.}
\label{tab:rewardbench}
\begin{tabular}{lcc}
\toprule
\textbf{Evaluator} & \textbf{ACC@Label} & \textbf{ACC@Critique} \\
\midrule
\texttt{Human}                   & 0.8355 & 0.7632 \\
\midrule
\texttt{gpt-5.4}                 & 0.4408 & 0.2039 \\
\texttt{claude-sonnet-4.6}       & 0.4474 & 0.1579 \\
\texttt{kimi-k2.5}               & 0.4473 & 0.1839 \\
\texttt{qwen3-max}               & 0.4276 & 0.2500 \\
\bottomrule
\end{tabular}
\end{table}

\paragraph{Results and Discussion.}
To establish a human performance baseline, we distribute each sample to a randomly and uniformly selected human annotator and obtain their independent judgments. The results reported in Tab.~\ref{tab:rewardbench} reveal that all LLM judges fall substantially short of human performance on both metrics. These findings demonstrate that automated LLM-as-a-Judge evaluation remains unreliable for social intelligence tasks of this nature, and underscore the necessity of human annotation in benchmark construction.

\section{Related Works}
\label{sec:related_work}

\paragraph{Social Intelligence.}
Human intelligence can be broadly categorized into logical intelligence and social intelligence~\citep{thorndike1920intelligence,goleman1996emotional,davis2011theory}. 
Social intelligence encompasses the understanding of one's own and others' mental states, such as emotions~\citep{goleman1996emotional,lane2010theory,grazzani2018relation}, motivations~\citep{braver2014mechanisms,simpson2015behavioral}, and values~\citep{leverage2010theory,apperly2012theory}; the capacity to empathize with vulnerable individuals~\citep{decety2010neurodevelopment,banissy2012inter}; the ability to respond appropriately to specific social contexts~\citep{hess2014role,laforest2021influence}; the competence to understand cultural differences~\citep{cole2003culture,han2015understanding}; the skills to resolve social conflicts~\citep{bryant1992conflict,overton2013conflict,adham2023conflict}; and the capability to persuade others through understanding their beliefs and knowledge states~\citep{friestad1994persuasion,falk2018persuasion}. 

\paragraph{Social Intelligence in LLMs.}
%
Currently, several efforts have attempted to evaluate social intelligence in LLMs; however, these evaluations remain fragmented, typically each focusing on one or a few isolated capability dimensions (summarized in Tab.~\ref{tab:comparison_of_social_benchmarks}). 
We categorize them as follows. 
(1) Some works focus only on surface-level character consistency, such as linguistic style, emotional tone, and knowledge~\citep{wang2024incharacter,zhou2025characterbench,lu2025rolemrc}. 
(2) Others evaluate capabilities such as emotion understanding~\citep{sabour2024emobench}, intention inference~\citep{li2025consintbench}, and false-belief reasoning~\citep{xu2024opentom}, which can be broadly categorized as Theory of Mind (ToM) reasoning~\citep{xu2024opentom,gu2024simpletom,shinoda2025tomato,wu2023hi}. 
(3) There are also higher-level benchmarks that assess the model's ability to produce contextually appropriate responses, such as emotional intelligence~\citep{eqbench3_repo_2025}, persuasion and negotiation~\citep{zhou2023sotopia}, and empathetic support in psychological counseling~\citep{zhang2025sentient}. Beyond these real-world scenarios, some works focus on virtual scenarios aimed at improving user experience~\citep{wang2025coser,xiang2025rmtbench}. For these benchmarks, the capabilities are typically connected to and shaped by the specific scenario.

\section{Conclusion}

We present SocialArena, a large-scale benchmark for systematically evaluating the social intelligence of LLMs. 
By structuring evaluation through a facts → mental states → behavior reasoning process, it enables a comprehensive assessment of social intelligence. 
Built upon the U‑TRIMF framework, SocialArena covers diverse social scenarios with varied interaction dynamics. 
Through a human-based arena-style evaluation protocol, SocialArena yields high-quality and reliable evaluation results.

\section*{Limitations}

We now discuss the limitations of SocialArena, which we anticipate addressing in future work. 

\paragraph{Potential Bias.}
Human preferences in conversational AI are influenced by individual backgrounds, communication styles, and subjective expectations. Our dataset is currently annotated by native English and Chinese speakers, which may introduce certain biases in preference judgments. 
To mitigate this limitation, we have developed and plan to release in the near future a public-facing platform \textbf{SocialArena-Platform}, similar to Chatbot Arena~\citep{zheng2023judging}, that enables broader community participation in data collection and evaluation. Such an approach would allow us to capture more diverse perspectives and interaction patterns, ultimately leading to more robust and generalizable models in future iterations of this work. We will release updated results once sufficient data has been collected. 
More details are provided in \S~\ref{appendix:social_arena_platform}. 

\paragraph{Multi-turn Dialogue.}
We adopt single-turn evaluation as our preliminary small-scale annotations revealed that multi-turn settings incur higher annotation costs, while single-turn scenarios sufficiently address most evaluation needs. 
Once sufficient single-turn evaluation data has been collected, we will enable multi-turn functionality on the platform and gather data to reduce the associated costs. 




\appendix

\clearpage
\bibliography{biblio}

@article{beaudoin2020systematic,
  title={Systematic review and inventory of theory of mind measures for young children},
  author={Beaudoin, Cindy and Leblanc, Elizabel and Gagner, Charlotte and Beauchamp, Miriam H},
  journal={Frontiers in psychology},
  volume={10},
  pages={2905},
  year={2020},
  publisher={Frontiers Media SA}
}

@article{zhang2025sentient,
  title={Sentient Agent as a Judge: Evaluating Higher-Order Social Cognition in Large Language Models},
  author={Zhang, Bang and Ma, Ruotian and Jiang, Qingxuan and Wang, Peisong and Chen, Jiaqi and Xie, Zheng and Chen, Xingyu and Wang, Yue and Ye, Fanghua and Li, Jian and others},
  journal={arXiv preprint arXiv:2505.02847},
  year={2025}
}

@article{grazzani2018relation,
  title={The relation between emotion understanding and theory of mind in children aged 3 to 8: The key role of language},
  author={Grazzani, Ilaria and Ornaghi, Veronica and Conte, Elisabetta and Pepe, Alessandro and Caprin, Claudia},
  journal={Frontiers in psychology},
  volume={9},
  pages={724},
  year={2018},
  publisher={Frontiers Media SA}
}

@article{lane2010theory,
  title={Theory of mind and emotion understanding predict moral development in early childhood},
  author={Lane, Jonathan D and Wellman, Henry M and Olson, Sheryl L and LaBounty, Jennifer and Kerr, David CR},
  journal={British Journal of Developmental Psychology},
  volume={28},
  number={4},
  pages={871--889},
  year={2010},
  publisher={Wiley Online Library}
}

@Inbook{simpson2015behavioral,
    author="Simpson, Eleanor H.
    and Balsam, Peter D.",
    editor="Simpson, Eleanor H.
    and Balsam, Peter D.",
    title="The Behavioral Neuroscience of Motivation: An Overview of Concepts, Measures, and Translational Applications",
    bookTitle="Behavioral Neuroscience of Motivation",
    year="2016",
    publisher="Springer International Publishing",
    address="Cham",
    pages="1--12",
    isbn="978-3-319-26935-1",
    doi="10.1007/7854_2015_402",
    url="https://doi.org/10.1007/7854_2015_402"
}

@article{braver2014mechanisms,
  title={Mechanisms of motivation--cognition interaction: challenges and opportunities},
  author={Braver, Todd S and Krug, Marie K and Chiew, Kimberly S and Kool, Wouter and Westbrook, J Andrew and Clement, Nathan J and Adcock, R Alison and Barch, Deanna M and Botvinick, Matthew M and Carver, Charles S and others},
  journal={Cognitive, affective, \& behavioral neuroscience},
  volume={14},
  number={2},
  pages={443--472},
  year={2014},
  publisher={Springer}
}

@article{decety2010neurodevelopment,
  title={The neurodevelopment of empathy in humans},
  author={Decety, Jean},
  journal={Developmental neuroscience},
  volume={32},
  number={4},
  pages={257--267},
  year={2010},
  publisher={S. Karger AG Basel, Switzerland}
}

@article{banissy2012inter,
  title={Inter-individual differences in empathy are reflected in human brain structure},
  author={Banissy, Michael J and Kanai, Ryota and Walsh, Vincent and Rees, Geraint},
  journal={Neuroimage},
  volume={62},
  number={3},
  pages={2034--2039},
  year={2012},
  publisher={Elsevier}
}

@book{leverage2010theory,
 URL = {http://www.jstor.org/stable/j.ctv15wxr83},
 edition = {DGO - Digital original},
 publisher = {Purdue University Press},
 title = {Theory of Mind and Literature},
 urldate = {2026-08-28},
 year = {2011}
}

@article{apperly2012theory,
  title={What is “theory of mind”? Concepts, cognitive processes and individual differences},
  author={Apperly, Ian A},
  journal={Quarterly journal of experimental psychology},
  volume={65},
  number={5},
  pages={825--839},
  year={2012},
  publisher={SAGE Publications Sage UK: London, England}
}

@incollection{hess2014role,
  title={The role of social context for the interpretation of emotional facial expressions},
  author={Hess, Ursula and Hareli, Shlomo},
  booktitle={Understanding facial expressions in communication: Cross-cultural and multidisciplinary perspectives},
  pages={119--141},
  year={2014},
  publisher={Springer}
}

@article{laforest2021influence,
  title={The influence of social context and social connection on visual perceptual processes},
  author={Laforest, Justin and MacGillivray, Megan and Lam, Melanie Y},
  journal={Acta Psychologica},
  volume={215},
  pages={103270},
  year={2021},
  publisher={Elsevier}
}

@article{cole2003culture,
  title={Culture and cognitive science},
  author={Cole, Michael},
  journal={Outlines. Critical Practice Studies},
  volume={5},
  number={1},
  pages={3--15},
  year={2003}
}

@article{han2015understanding,
  title={Understanding cultural differences in human behavior: a cultural neuroscience approach},
  author={Han, Shihui},
  journal={Current opinion in behavioral sciences},
  volume={3},
  pages={68--72},
  year={2015},
  publisher={Elsevier}
}

@article{bryant1992conflict,
  title={Conflict resolution strategies in relation to children's peer relations},
  author={Bryant, Brenda K},
  journal={Journal of Applied Developmental Psychology},
  volume={13},
  number={1},
  pages={35--50},
  year={1992},
  publisher={Elsevier}
}

@article{adham2023conflict,
  title={Conflict resolution in team: Analyzing the of conflicts and best skills for resolution},
  author={Adham, Tawfig Khidir Ibnouf},
  journal={Scholars Journal of Engineering and Technology},
  volume={11},
  number={08},
  pages={152--162},
  year={2023}
}

@article{overton2013conflict,
  title={Conflict management: difficult conversations with difficult people},
  author={Overton, Amy R and Lowry, Ann C},
  journal={Clinics in colon and rectal surgery},
  volume={26},
  number={04},
  pages={259--264},
  year={2013},
  publisher={Thieme Medical Publishers}
}

@article{falk2018persuasion,
  title={Persuasion, influence, and value: Perspectives from communication and social neuroscience},
  author={Falk, Emily and Scholz, Christin},
  journal={Annual review of psychology},
  volume={69},
  number={1},
  pages={329--356},
  year={2018},
  publisher={Annual Reviews}
}

@article{friestad1994persuasion,
  title={The persuasion knowledge model: How people cope with persuasion attempts},
  author={Friestad, Marian and Wright, Peter},
  journal={Journal of consumer research},
  volume={21},
  number={1},
  pages={1--31},
  year={1994},
  publisher={The University of Chicago Press}
}

@article{zhou2023sotopia,
  title={Sotopia: Interactive evaluation for social intelligence in language agents},
  author={Zhou, Xuhui and Zhu, Hao and Mathur, Leena and Zhang, Ruohong and Yu, Haofei and Qi, Zhengyang and Morency, Louis-Philippe and Bisk, Yonatan and Fried, Daniel and Neubig, Graham and others},
  journal={arXiv preprint arXiv:2310.11667},
  year={2023}
}

@inproceedings{sabour2024emobench,
  title={Emobench: Evaluating the emotional intelligence of large language models},
  author={Sabour, Sahand and Liu, Siyang and Zhang, Zheyuan and Liu, June and Zhou, Jinfeng and Sunaryo, Alvionna and Lee, Tatia and Mihalcea, Rada and Huang, Minlie},
  booktitle={Proceedings of the 62nd Annual Meeting of the Association for Computational Linguistics (Volume 1: Long Papers)},
  pages={5986--6004},
  year={2024}
}

@inproceedings{chen2024socialbench,
  title={Socialbench: Sociality evaluation of role-playing conversational agents},
  author={Chen, Hongzhan and Chen, Hehong and Yan, Ming and Xu, Wenshen and Xing, Gao and Shen, Weizhou and Quan, Xiaojun and Li, Chenliang and Zhang, Ji and Huang, Fei},
  booktitle={Findings of the Association for Computational Linguistics: ACL 2024},
  pages={2108--2126},
  year={2024}
}

@article{huang2025si,
  title={SI-Bench: Benchmarking Social Intelligence of Large Language Models in Human-to-Human Conversations},
  author={Huang, Shuai and Zhao, Wenxuan and Gao, Jun},
  journal={arXiv preprint arXiv:2510.23182},
  year={2025}
}

@inproceedings{shinoda2025tomato,
  title={Tomato: Verbalizing the mental states of role-playing llms for benchmarking theory of mind},
  author={Shinoda, Kazutoshi and Hojo, Nobukatsu and Nishida, Kyosuke and Mizuno, Saki and Suzuki, Keita and Masumura, Ryo and Sugiyama, Hiroaki and Saito, Kuniko},
  booktitle={Proceedings of the AAAI Conference on Artificial Intelligence},
  volume={39},
  number={2},
  pages={1520--1528},
  year={2025}
}

@article{gu2024simpletom,
  title={Simpletom: Exposing the gap between explicit tom inference and implicit tom application in llms},
  author={Gu, Yuling and Tafjord, Oyvind and Kim, Hyunwoo and Moore, Jared and Bras, Ronan Le and Clark, Peter and Choi, Yejin},
  journal={arXiv preprint arXiv:2410.13648},
  year={2024}
}

@article{xu2024opentom,
  title={OpenToM: A comprehensive benchmark for evaluating theory-of-mind reasoning capabilities of large language models},
  author={Xu, Hainiu and Zhao, Runcong and Zhu, Lixing and Du, Jinhua and He, Yulan},
  journal={arXiv preprint arXiv:2402.06044},
  year={2024}
}

@inproceedings{wu2023hi,
  title={Hi-tom: A benchmark for evaluating higher-order theory of mind reasoning in large language models},
  author={Wu, Yufan and He, Yinghui and Jia, Yilin and Mihalcea, Rada and Chen, Yulong and Deng, Naihao},
  booktitle={Findings of the Association for Computational Linguistics: EMNLP 2023},
  pages={10691--10706},
  year={2023}
}

@article{xiang2025rmtbench,
  title={RMTBench: Benchmarking LLMs Through Multi-Turn User-Centric Role-Playing},
  author={Xiang, Hao and Tang, Tianyi and Su, Yang and Yu, Bowen and Yang, An and Huang, Fei and Zhang, Yichang and Lu, Yaojie and Lin, Hongyu and Han, Xianpei and others},
  journal={arXiv preprint arXiv:2507.20352},
  year={2025}
}

@article{wang2025coser,
  title={Coser: Coordinating llm-based persona simulation of established roles},
  author={Wang, Xintao and Wang, Heng and Zhang, Yifei and Yuan, Xinfeng and Xu, Rui and Huang, Jen-tse and Yuan, Siyu and Guo, Haoran and Chen, Jiangjie and Zhou, Shuchang and others},
  journal={arXiv preprint arXiv:2502.09082},
  year={2025}
}

@article{lu2025rolemrc,
  title={Rolemrc: A fine-grained composite benchmark for role-playing and instruction-following},
  author={Lu, Junru and Li, Jiazheng and Shen, Guodong and Gui, Lin and An, Siyu and He, Yulan and Yin, Di and Sun, Xing},
  journal={arXiv preprint arXiv:2502.11387},
  year={2025}
}

@inproceedings{zhou2025characterbench,
  title={CharacterBench: Benchmarking Character Customization of Large Language Models},
  author={Zhou, Jinfeng and Huang, Yongkang and Wen, Bosi and Bi, Guanqun and Chen, Yuxuan and Ke, Pei and Chen, Zhuang and Xiao, Xiyao and Peng, Libiao and Tang, Kuntian and others},
  booktitle={Proceedings of the AAAI Conference on Artificial Intelligence},
  volume={39},
  number={24},
  pages={26101--26110},
  year={2025}
}

@article{xu2026adamarp,
  title={AdaMARP: An Adaptive Multi-Agent Interaction Framework for General Immersive Role-Playing},
  author={Xu, Zhenhua and Chen, Dongsheng and Wang, Shuo and Li, Jian and Wang, Chengjie and Han, Meng and Wang, Yabiao},
  journal={arXiv preprint arXiv:2601.11007},
  year={2026}
}

@article{sap2019socialiqa,
  title={Socialiqa: Commonsense reasoning about social interactions},
  author={Sap, Maarten and Rashkin, Hannah and Chen, Derek and LeBras, Ronan and Choi, Yejin},
  journal={arXiv preprint arXiv:1904.09728},
  year={2019}
}

@misc{eqbench3_repo_2025,
  author       = {Samuel J. Paech},
  title        = {EQ-Bench 3: Emotional Intelligence Benchmark},
  year         = {2025},
  howpublished = {\url{https://github.com/EQ-bench/eqbench3}}
}

@inbook{davis2011theory, 
    place={Cambridge}, 
    series={Cambridge Handbooks in Psychology}, 
    title={The Theory of Multiple Intelligences}, booktitle={The Cambridge Handbook of Intelligence}, 
    publisher={Cambridge University Press}, 
    author={Davis, Katie and Christodoulou, Joanna and Seider, Scott and Gardner, Howard}, 
    editor={Sternberg, Robert J. and Kaufman, Scott BarryEditors}, 
    year={2011}, 
    pages={485–503}, 
    collection={Cambridge Handbooks in Psychology}
}

@article{thorndike1920intelligence,
  author  = {Thorndike, E. L.},
  title   = {Intelligence and its uses},
  journal = {Harper's Magazine},
  year    = {1920},
  volume  = {140},
  pages   = {227--235}
}

@article{goleman1996emotional,
  title={Emotional intelligence. Why it can matter more than IQ.},
  author={Goleman, Daniel},
  journal={Learning},
  volume={24},
  number={6},
  pages={49--50},
  year={1996},
  publisher={ERIC}
}

@article{guo2025deepseek,
  title={Deepseek-r1: Incentivizing reasoning capability in llms via reinforcement learning},
  author={Guo, Daya and Yang, Dejian and Zhang, Haowei and Song, Junxiao and Zhang, Ruoyu and Xu, Runxin and Zhu, Qihao and Ma, Shirong and Wang, Peiyi and Bi, Xiao and others},
  journal={arXiv preprint arXiv:2501.12948},
  year={2025}
}

@article{fiske1992four,
  title={The four elementary forms of sociality: framework for a unified theory of social relations.},
  author={Fiske, Alan P},
  journal={Psychological review},
  volume={99},
  number={4},
  year={1992},
  publisher={American Psychological Association}
}

@article{kluver1999topology,
  title={Topology, metric and dynamics of social systems},
  author={Kl{\"u}ver, J{\"u}rgen and Schmidt, J{\"o}rn and others},
  journal={Journal of Artificial Societies and Social Simulation},
  volume={2},
  number={3},
  pages={7},
  year={1999},
  publisher={Journal of Artificial Societies and Social Simulation}
}

@misc{mcleod2007maslow,
  author       = {{Wikipedia contributors}},
  title        = {Maslow's Hierarchy of Needs},
  year         = {2026},
  howpublished = {\url{https://en.wikipedia.org/wiki/Maslow%27s_hierarchy_of_needs}},
  note         = {Accessed: 2026-08-28}
}

@article{hollinger1982formal,
  title={Formal and informal social controls of employee deviance},
  author={Hollinger, Richard C and Clark, John P},
  journal={Sociological quarterly},
  volume={23},
  number={3},
  pages={333--343},
  year={1982},
  publisher={Wiley Online Library}
}

@book{fudenberg1991game,
  title={Game theory},
  author={Fudenberg, Drew and Tirole, Jean},
  year={1991},
  publisher={MIT press}
}

@article{wang2026reward,
  title={Reward Modeling from Natural Language Human Feedback},
  author={Wang, Zongqi and Wang, Rui and Wu, Yuchuan and Yu, Yiyao and Zhang, Pinyi and Sun, Shaoning and Yang, Yujiu and Li, Yongbin},
  journal={arXiv preprint arXiv:2601.07349},
  year={2026}
}

@book{elo1966uscf,
  title={The USCF Rating System: Its Development, Theory, and Applications},
  author={Elo, A.E.},
  url={https://books.google.com.sg/books?id=onUazQEACAAJ},
  year={1966},
  publisher={United States Chess Federation}
}

@misc{qwen3technicalreport,
  title={Qwen3 Technical Report}, 
  author={Qwen Team},
  year={2025},
  eprint={2505.09388},
  archivePrefix={arXiv},
  primaryClass={cs.CL},
  url={https://arxiv.org/abs/2505.09388}, 
}

@inproceedings{wang2024incharacter,
  title={Incharacter: Evaluating personality fidelity in role-playing agents through psychological interviews},
  author={Wang, Xintao and Xiao, Yunze and Huang, Jen-tse and Yuan, Siyu and Xu, Rui and Guo, Haoran and Tu, Quan and Fei, Yaying and Leng, Ziang and Wang, Wei and others},
  booktitle={Proceedings of the 62nd annual meeting of the association for computational linguistics (volume 1: Long papers)},
  pages={1840--1873},
  year={2024}
}

@inproceedings{park2023generative,
  title={Generative agents: Interactive simulacra of human behavior},
  author={Park, Joon Sung and O'Brien, Joseph and Cai, Carrie Jun and Morris, Meredith Ringel and Liang, Percy and Bernstein, Michael S},
  booktitle={Proceedings of the 36th annual acm symposium on user interface software and technology},
  pages={1--22},
  year={2023}
}

@inproceedings{markel2023gpteach,
  title={Gpteach: Interactive ta training with gpt-based students},
  author={Markel, Julia M and Opferman, Steven G and Landay, James A and Piech, Chris},
  booktitle={Proceedings of the tenth acm conference on learning@ scale},
  pages={226--236},
  year={2023}
}

@article{zheng2023judging,
  title={Judging llm-as-a-judge with mt-bench and chatbot arena},
  author={Zheng, Lianmin and Chiang, Wei-Lin and Sheng, Ying and Zhuang, Siyuan and Wu, Zhanghao and Zhuang, Yonghao and Lin, Zi and Li, Zhuohan and Li, Dacheng and Xing, Eric and others},
  journal={Advances in neural information processing systems},
  volume={36},
  pages={46595--46623},
  year={2023}
}

@article{li2025consintbench,
  title={ConsintBench: Evaluating Language Models on Real-World Consumer Intent Understanding},
  author={Li, Xiaozhe and Lyu, TianYi and Yang, Siyi and Gong, Yuxi and Yang, Yizhao and Huang, Jinxuan and Zhang, Ligao and Huang, Zhuoyi and Liu, Qingwen},
  journal={arXiv preprint arXiv:2510.13499},
  year={2025}
}

@inproceedings{zhao2024esc,
  title={Esc-eval: Evaluating emotion support conversations in large language models},
  author={Zhao, Haiquan and Li, Lingyu and Chen, Shisong and Kong, Shuqi and Wang, Jiaan and Huang, Kexin and Gu, Tianle and Wang, Yixu and Wang, Jian and Dandan, Liang and others},
  booktitle={Proceedings of the 2024 Conference on Empirical Methods in Natural Language Processing},
  pages={15785--15810},
  year={2024}
}

@misc{gutenberg,
  title        = {Project {Gutenberg}},
  howpublished = {\url{https://www.gutenberg.org}},
  author  = {{Project Gutenberg}},
  urldate = {2026-08-28}
}

@article{shao2025deepseekmath,
  title={Deepseekmath-v2: Towards self-verifiable mathematical reasoning},
  author={Shao, Zhihong and Luo, Yuxiang and Lu, Chengda and Ren, ZZ and Hu, Jiewen and Ye, Tian and Gou, Zhibin and Ma, Shirong and Zhang, Xiaokang},
  journal={arXiv preprint arXiv:2511.22570},
  year={2025}
}

@article{jimenez2023swe,
  title={Swe-bench: Can language models resolve real-world github issues?},
  author={Jimenez, Carlos E and Yang, John and Wettig, Alexander and Yao, Shunyu and Pei, Kexin and Press, Ofir and Narasimhan, Karthik},
  journal={arXiv preprint arXiv:2310.06770},
  year={2023}
}

@article{barres2025tau,
  title={$tau^2$-Bench: Evaluating Conversational Agents in a Dual-Control Environment},
  author={Barres, Victor and Dong, Honghua and Ray, Soham and Si, Xujie and Narasimhan, Karthik},
  journal={arXiv preprint arXiv:2506.07982},
  year={2025}
}

@article{zeng2026glm,
  title={Glm-5: from vibe coding to agentic engineering},
  author={Zeng, Aohan and Lv, Xin and Hou, Zhenyu and Du, Zhengxiao and Zheng, Qinkai and Chen, Bin and Yin, Da and Ge, Chendi and Huang, Chenghua and Xie, Chengxing and others},
  journal={arXiv preprint arXiv:2602.15763},
  year={2026}
}

@misc{qwen35blog,
    title = {Qwen3.5: Accelerating Productivity with Native Multimodal Agents},
    url = {https://qwen.ai/blog?id=qwen3.5},
    author = {Qwen Team},
    month = {February},
    year = {2026}
}
\bibliographystyle{colm2024_conference}

\newpage
\appendix

\section{Discussion}
\label{appendix:discussion}

\subsection{Impact Statement}
\label{appendix:impact_statement}

\paragraph{Positive Aspects.} Our research will promote the development of humanitarian care in the LLM field. It can significantly advance research on reward modeling and RLHF in unverifiable domains. The academic community can leverage this to obtain high-quality data and develop more powerful character models with social intelligence. 

\paragraph{Negative Aspects.} The social scenarios on SocialArena may themselves contain unhealthy or unsafe content (e.g., illegal or unethical material), which could negatively impact academic research if used for model training or evaluation. To address this, our platform implements strict filtering mechanisms to exclude such content from all displayed scenarios and leaderboard updates. 


\section{SocialArena-Platform}
\label{appendix:social_arena_platform}

\paragraph{Overview.}
To facilitate continuous data collection and broader community engagement, we introduce \textbf{SocialArena-Platform}, a publicly accessible online platform designed to complement and extend the SocialArena benchmark. 

\paragraph{Motivation.}
The motivation behind this platform is twofold. 
First, human preferences in social interactions are inherently diverse, and a static dataset annotated by a limited group of annotators may not fully capture this diversity. 
Second, real-world social dilemmas are dynamic and ever-evolving, necessitating a living platform that can continuously collect fresh scenarios and preference judgments beyond the scope of our initial benchmark. 

\paragraph{Platform Design.}
The platform is centered around real-world social interaction. Users can post the social dilemmas they actually encounter in their daily lives, receive responses from state-of-the-art LLMs as well as fellow community members, and ultimately find resolutions to their challenges. Meanwhile, community upvotes and comments serve as a natural and scalable evaluation mechanism for assessing model capabilities in social intelligence. 

\paragraph{User Engagement.}
To attract a broad and diverse user base, we focus on delivering immediate practical value to general users. The platform offers free access to multiple state-of-the-art LLMs, allowing users to directly experience and compare model responses to real social dilemmas at no cost. Moreover, the platform hosts a rich repository of diverse social scenarios along with corresponding model responses, which users can freely browse for reference and inspiration when navigating their own interpersonal challenges. We believe that this combination of free model access and rich reference content will naturally draw users who are genuinely interested in seeking advice and exploring how AI can assist in complex social situations. 

\paragraph{Data Quality Control.}
To ensure the reliability of annotations collected in the wild, we adopt a multi-layered filtering strategy. 
First, we filter out prompts that do not correspond to social interaction tasks, such as coding or mathematical queries, to ensure that all collected scenarios fall within the intended domain. 
Second, we apply the same scenario qualification criteria introduced in Section~\ref{sec:dataset_curation_pipeline} to remove low-quality or ill-formed scenarios that fail to meet our benchmark standards. 
Third, we adopt the anomalous IP detection rules from Chatbot Arena~\citep{zheng2023judging} to identify and exclude votes originating from suspicious or duplicate sources, thereby safeguarding the integrity of the collected preference judgments. 

Through these efforts, we hope that SocialArena-Platform can serve as a sustainable and community-driven resource that genuinely contributes to human well-being and advances the development of socially intelligent AI systems.


\section{Details of Benchmark Construction}
\label{appendix:benchmark_construction_details}

\subsection{U-TRIMF Framework}

\begin{table*}[h!]
\centering
\small
\caption{The U-TRIMF Framework for Base Scenario Construction.}
\begin{tabular}{p{2.3cm}p{4.0cm}p{8.5cm}}
\toprule
\textbf{Dimension} & \textbf{Category} & \textbf{Examples} \\
\midrule
\rowcolor{gray!15}
\multicolumn{3}{l}{\textit{Basic Element 1: User Presence}} \\
\midrule
\multirow{2}{2.3cm}{\textbf{U}ser Presence} & U1: Real-world & LLM acts as a character governed by real-world constraints; no user present \\
 & U2: User-centered & LLM serves a user's specific needs; user present and drives scenario evolution \\
\midrule
\rowcolor{gray!15}
\multicolumn{3}{l}{\textit{Basic Element 2: Characters}} \\
\midrule
\multirow{3}{2.3cm}{\textbf{T}opological Relation} & \sout{T1: Individual} & \textit{(Excluded: a solitary individual cannot constitute social interaction)} \\
 & T2: Dyad & Couple, doctor/patient, teacher/student, buyer/seller \\
 & T3: Group & Team members, family members, book club, colleagues \\
\midrule
\multirow{5}{2.3cm}{\textbf{R}elational Model} & R1: Communal Sharing & Mother/child, romantic partners, parent/offspring \\
 & R2: Authority Ranking & CEO/employee, advisor/graduate student, officer/soldier \\
 & R3: Equality Matching & Colleagues, classmates, neighbors, carpool partners, friends \\
 & R4: Market Pricing & Consumer/merchant, landlord/tenant, employer/employee \\
 & R5: Null & Strangers, passersby, pedestrians \\
\midrule
\rowcolor{gray!15}
\multicolumn{3}{l}{\textit{Basic Element 3: Events}} \\
\midrule
\multirow{3}{2.3cm}{\textbf{I}nterest Structure} & I1: Zero-Sum & Bidding for projects, election campaigns, inheritance disputes \\
 & I2: Negative-Sum & Price wars, divorce litigation, trade wars, factional conflicts \\
 & I3: Positive-Sum & Strategic partnerships, joint ventures, community building \\
\midrule
\multirow{5}{2.3cm}{\textbf{M}otivational Type} & M1: Physiological & Eating (hungry), drinking (thirsty), sleeping (tired) \\
 & M2: Safety & Buying insurance, saving money, installing security systems \\
 & M3: Belonging & Family reunion, joining communities, team building \\
 & M4: Esteem & Competing for awards, conference presentations \\
 & M5: Self-actualization & Teaching volunteering, studying art, pursuing PhD \\
\midrule
\multirow{2}{2.3cm}{\textbf{F}ormality} & F1: Informal & Value exploration, stranger storytelling, airport deep talks \\
 & F2: Formal & Psychological therapy, estate distribution, board decisions \\
\bottomrule
\end{tabular}
\label{tab:trimf}
\end{table*}

The U-TRIMF framework is a structured taxonomy we designed to systematically cover the space of basic social contexts. It decomposes a social scenario into three fundamental elements---\textbf{User Presence}, \textbf{Characters}, and \textbf{Events}---and characterizes each element along multiple theoretically grounded dimensions. Specifically, user presence is described by the \textbf{U}ser Presence dimension; characters are described by their \textbf{T}opological Relationships and \textbf{R}elational Models; events are characterized by their \textbf{I}nterest Structure, \textbf{M}otivational Types, and \textbf{F}ormality level. Together, these six dimensions give rise to the acronym \textbf{U-TRIMF}. 

\paragraph{User Presence.} The \textit{User Presence} dimension captures whether a human user exists in the scenario, which fundamentally shapes the LLM's role and responsibilities~. In \textbf{U1 (Real-world)} scenarios, no user is present: the LLM acts as a character embedded in and governed by the real world~\citep{zhang2025sentient, huang2025si}. In \textbf{U2 (User-centered)} scenarios, a user is present: the LLM is tasked with serving user-specific needs, and accordingly gains the authority to influence how the scenario evolves~\citep{wang2025coser,xiang2025rmtbench}. 

\paragraph{Characters.} The \textit{Topological Relationship} dimension captures the structural configuration of participants in a social scene. We adopt two active categories: a symmetric dyad (T2) and a symmetric group (T3). Importantly, T1 (a single individual alone) is \textbf{excluded} from our benchmark construction, as a solitary person cannot constitute a social interaction by definition. All scenarios in our benchmark therefore involve at least two participants (T2 or T3). The \textit{Relational Model} dimension, grounded in Fiske's~\citeyearpar{fiske1992four} influential framework, encodes the qualitative nature of the relationship between participants, ranging from intimate communal sharing (R1) to purely transactional market pricing (R4) and even null/asocial ties between strangers (R5). 

\paragraph{Events.} The \textit{Interest Structure} dimension (I) captures the game-theoretic structure of the situation~\citep{fudenberg1991game}, covering zero-sum, negative-sum, and positive-sum interactions. The \textit{Motivational Type} dimension (M) is adapted from Maslow's hierarchy of needs~\citeyearpar{mcleod2007maslow}, spanning physiological needs (M1) through self-actualization (M5). Finally, the \textit{Formality} dimension (F) distinguishes between informal, everyday interactions (F1) and formally structured, institutionalized settings (F2). 

A complete description of all categories with illustrative examples is provided in Table~\ref{tab:trimf}.

\subsection{Explanation of Keyword-Guided Scenario Synthesis}
\label{appendix:keyword_guided}

\paragraph{Why keyword guidance?}
A naive approach to scenario generation---prompting a language model directly with a U-TRIMF category combination---produces highly repetitive outputs, as large language models tend to converge on the most prototypical exemplar of any given category. To mitigate this degeneracy, we adopt a \textbf{keyword-guided synthesis} strategy: each generation call is conditioned on 10 thematic keywords that seed the narrative content after the U-TRIMF constraints are applied.

\paragraph{Keyword sourcing and design.}
Keywords are extracted from randomly sampled 10,000 chapters from 10,000 books in \textbf{Project Gutenberg~\citep{gutenberg}} (with professional/technical books filtered out). Each collection of 10 keywords is drawn from a single chapter of a book to ensure that the 10 keywords remain thematically coherent rather than semantically disparate.

\paragraph{Keyword guidance does not introduce bias.}
We argue here that keyword guidance does not introduce bias into scenario synthesis. First, keywords serve only as soft guidance rather than hard constraints: the model is instructed to draw inspiration from them rather than incorporate all of them verbatim. Second, the pre-established U-TRIMF template acts as a structural scaffold that ensures comprehensive coverage of the scenario space, rather than the keywords doing so. Finally, all generated instances are manually verified to ensure overall data quality.

\subsection{Explanation of Quality Filtering}
\label{appendix:quality_filtering}

Raw scenarios generated by the language model are passed through a two-stage quality filtering pipeline: (1) a set of explicit \textbf{rule-based filters} that automatically reject scenarios violating pre-specified constraints, implemented via \texttt{Qwen3-235B-A22B-Instruct-2507}, followed by (2) a \textbf{human review stage} in which annotators make holistic judgments about scenario suitability. 

\subsubsection{Rule-Based Filters}

The rules fall into four thematic groups.

\paragraph{Group A: Content Safety \& Ethical Boundaries.}
These rules reject scenarios whose content would make them inappropriate or distressing to present to participants, regardless of their social interest.
\begin{itemize}
    \item \textbf{A1 --- No life-threatening content.} Scenarios must not involve serious illness, terminal diagnosis, suicide, life-threatening danger, or near-death experiences. \textit{Exception:} a scenario in which one party provides positive emotional support to the other (e.g., comforting someone anxious about a routine medical check-up) is permitted.
    \item \textbf{A2 --- No serious criminal conduct.} Scenarios must not involve murder, sexual assault, drug trafficking, kidnapping, or comparable crimes.
    \item \textbf{A3 --- No extreme trauma.} Scenarios must not depict childhood abuse, abandonment, or severe violence as central narrative elements.
    \item \textbf{A4 --- No illegal events.} Scenarios must not involve legally prohibited acts such as marital infidelity that constitutes fraud, embezzlement, or financial misconduct.
    \item \textbf{A5 --- No major betrayal.} Scenarios must not hinge on acts of serious interpersonal betrayal (e.g., informing on a close friend to authorities).
    \item \textbf{A6 --- No irreversible catastrophic loss.} Scenarios must not involve bankruptcy, complete reputational destruction, or equivalent permanent, irrecoverable harms as a primary plot element.
\end{itemize}

\paragraph{Group B: Scenario Complexity \& Conflict Validity.}
These rules ensure that each scenario presents a genuinely non-trivial social situation.
\begin{itemize}
    \item \textbf{B1 --- No trivial everyday disputes.} Scenarios that lack any meaningful complexity---where any resolution proposed by either party would be equally acceptable at negligible cost---are excluded.
    \item \textbf{B2 --- No extreme or forced dilemmas.} Scenarios must not place characters in all-or-nothing situations requiring one party to suffer a severe, unavoidable harm. The tension should be realistic and proportionate.
    \item \textbf{B3 --- Single-event focus.} Each scenario must revolve around a single, coherent event or situation. Scenarios that superimpose multiple independent conflicts are excluded.
    \item \textbf{B4 --- Objective-basis requirement for conflict.} If a scenario contains a conflict, it must arise from objective circumstances (e.g., a resource constraint, an impending deadline, a genuine incompatibility of needs) rather than from artificially or inexplicably manufactured antagonism.
    \item \textbf{B5 --- Balanced conflict weight.} When a scenario features opposing interests, both sides must carry comparable moral, practical, or emotional weight. Asymmetric conflicts---where one side concerns safety or fundamental rights and the other concerns minor personal preferences---are excluded. Example of a rejected pair: \textit{child safety on a balcony} vs.\ \textit{aesthetic preference for an open railing}.
\end{itemize}

\paragraph{Group C: Structural \& Narrative Requirements.}
These rules enforce properties of the scenario's construction that are necessary for its use in the benchmark.
\begin{itemize}
    \item \textbf{C1 --- Appropriate conversational entry point.} The opening dialogue included in each scenario must end at a moment that naturally invites a continuation---a point of genuine tension, an unanswered question, or a decision hanging in the air---so that a model assigned to role-play a character has meaningful space to respond.
    \item \textbf{C2 --- Unbroken narrative logic.} The causal chain within the scenario must be internally coherent. Scenarios in which a character's reaction is logically incomprehensible given the stated trigger are rejected. \textit{Example of a rejected scenario:} one character cuts their hair short; the other accuses them of ``being the villain''---the logical connection is missing.
\end{itemize}

\paragraph{Group D: Conflict Substance Requirements.}
These rules guard against scenarios whose conflict is only superficially social.
\begin{itemize}
    \item \textbf{D1 --- Psychological or interest-based conflict required.} \textit{(If this is a real social scenario; virtual/fictional scenarios are exempt from this requirement.)} The conflict must engage genuine psychological or interpersonal stakes---emotional needs, relationship maintenance, value disagreements, identity concerns, material interests, or the need to persuade another person. Purely technical or procedural trade-offs are excluded. \textit{Example of a rejected scenario:} two colleagues debating which document structure to use for a report, or which software tool is more efficient.

    \item \textbf{D2 --- Sufficient emotional weight.} \textit{(If this is a real social scenario; virtual/fictional scenarios are exempt from this requirement.)} The conflict must touch something the characters genuinely care about: a significant relationship, a core value, a meaningful identity, or a decision with real consequences for their lives. Trivial, surface-level disagreements are excluded. \textit{Examples of rejected scenarios:} neighbors casually discussing whether to remove an interior door; coworkers debating which restaurant to order lunch from.

    \item \textbf{D3 --- Immediate action pressure required.} \textit{(If this is a real social scenario; virtual/fictional scenarios are exempt from this requirement.)} The scenario must create a sense of urgency---someone is waiting for a response, a decision must be made on the spot, or a time window is about to close. Scenarios framed as open-ended reflection, long-term planning, or interruptible deliberation are excluded. \textit{Examples of rejected scenarios:} a character pondering life goals after a routine health check-up; a character receiving a message and drifting into private rumination.

    \item \textbf{D4 --- No ``free exit'' situations.} \textit{(If this is a real social scenario; virtual/fictional scenarios are exempt from this requirement.)} Characters must have a genuine stake in the interaction such that walking away is not a trivially costless option. Scenarios in which a character has no real reason to remain engaged yet is portrayed as conflicted are rejected. \textit{Example of a rejected scenario:} a tenant who has no contractual obligation to a property manager deliberating at length over whether to sign a discretionary form, when they can simply leave.
\end{itemize}

\subsubsection{Human Review}

Scenarios that pass all rule-based filters are subsequently reviewed by human annotators. Each annotator reads the full scenario---including the character descriptions, situational background, and opening dialogue---and makes a holistic judgment along the following dimensions:

\begin{enumerate}
    \item \textbf{Overall plausibility.} If this is a real social scenario, does it feel like something that could realistically happen between real people? Does the emotional and behavioral logic of the characters hold up under scrutiny? If this is a virtual social scenario, is it actually something that users would use or experience?
    \item \textbf{Balance and fairness of the conflict.} Would a reasonable person find both sides of the conflict understandable and legitimate, even if asymmetric in some minor respect?
    \item \textbf{Role-play affordance.} Does the scenario provide a natural, productive entry point for a model to inhabit one of the characters? Is the situation open enough to allow multiple plausible responses, yet constrained enough to require genuine social reasoning?
    \item \textbf{Freedom from subtle rule violations.} Does the scenario pass the spirit---not merely the letter---of the rules above? Annotators flag cases that technically satisfy each individual rule but collectively feel implausible, gratuitous, or inappropriate.
\end{enumerate}

Scenarios are retained only if the annotator judges them as satisfactory on all four dimensions. Disagreements between annotators are resolved through discussion, and borderline cases are discarded to maintain a conservative quality threshold.


\section{Details of Evaluation Protocol}
\label{appendix:evaluation_protocol}

\subsection{Annotation Guidelines}
\label{appendix:annotation_guidelines}

\begin{table*}[h!]
\centering
\small
\caption{Annotation guidelines provided to annotators.}
\label{tab:annotation_guidelines}
\begin{tabular}{p{3.0cm} p{11.5cm}}
\toprule
\textbf{Item} & \textbf{Guideline} \\
\midrule
\textbf{Task Objective} & Judge which of two model responses (A or B) better reflects social intelligence. Label using \textbf{A} (A better), \textbf{B} (B better), \textbf{T} (both good), or \textbf{G} (both poor). \\
\midrule
\textbf{Thinking Evaluation} & Judge which response provides a better analysis of each mental state dimension. Evaluation principles: \textbf{Correctness first}: if a response's thinking contains an objectively wrong inference about any mental state (intent, emotion, belief, etc.), it loses regardless of other qualities. \textbf{Utility second}: a correct analysis counts as better only if it meaningfully promotes a better behavior; a correct but redundant analysis is scored T. \\
\midrule
\textbf{Behavior Evaluation} & In general, \textbf{prefer responses that demonstrate higher social intelligence}, such as being more emotionally intelligent, empathetic, persuasive, etc. However, there are some exceptions: a. \textbf{Long-term interests take priority:} If making a compromise would make oneself appear weak, leading to short-term problem resolution but long-term negative consequences, then compromising is not acceptable. b. \textbf{Maintain moral boundaries:} If there is a conflict between illegal behavior and personal interests, choosing personal interest is not acceptable. If the other party has committed a moral violation — such as cheating, stealing, etc. — be careful \textbf{not} to empathize with them in such cases. c. \textbf{Role consistency:} If a character is explicitly described as someone who is "psychologically traumatized," "extremely angry," "timid and fearful," or "very hesitant," care must be taken \textbf{not to violate the character's established traits}. For example, if a character has psychological trauma, they should not immediately comfort themselves and recover quickly at the beginning. \\
\midrule
\textbf{Factual Grounding} & Any content in a response that cannot be found in the scenario background---and cannot be naturally and reasonably inferred---counts as a factual error and penalizes that response. \\
\midrule
\textbf{Conflict Weight} & When only one response covers a particular mental state dimension, it wins on that dimension only if covering it produces a meaningfully better behavior; otherwise score T. When neither response covers a dimension, leave that field blank. \\
\midrule
\textbf{Comment Requirements} & For every dimension, identify the specific strengths and weaknesses of both A and B. Abstract descriptors (e.g., ``more thorough'', ``clearer'') are not permitted; comments should cite concrete content from the responses wherever possible. \\
\bottomrule
\end{tabular}
\end{table*}

Annotators were provided with two forms of guidance prior to beginning the annotation task. First, they received a structured guidelines document covering the foundational rules of the annotation protocol, summarized in Table~\ref{tab:annotation_guidelines}. Second, they were given a set of worked examples---including full scenario contexts, paired model responses, and expert-written reference annotations---to calibrate their understanding of each dimension and the ABTG label scheme before handling live data. 

\subsection{Annotator Recruitment and Quality Control}
\label{appendix:annotator_ethics}

Annotations were recruited provided through third‑party annotation vendors. The annotator pool consisted of a mix of native English speakers and native Chinese speakers in roughly equal proportion. 

All annotators were informed about the nature of the data and the intended research use of their annotations through the vendor’s standard onboarding and consent procedures. No personally identifiable information about annotators was shared with the research team beyond what was necessary for managing the annotation workflow through the vendor. 

To maintain annotation quality, we implemented a batch-level quality control process. For each annotation batch, every annotator had a small subset of their submitted items independently reviewed by internal expert reviewers. Annotators whose quality pass rate fell below 75\% for a batch had their corresponding annotations removed from the dataset. 
Additionally, these annotators received targeted feedback and further training to help them improve their performance in subsequent batches. 

\subsection{Details of Theory of Mind (ToM)}
\label{appendix:tom_by_mental_states}

We adopt the taxonomy of mental state categories from \citet{beaudoin2020systematic}, which provides a comprehensive and theoretically grounded classification of the mental states implicated in Theory of Mind reasoning. Specifically, we operationalize ToM across the following seven categories, summarized in Table~\ref{tab:tom_mental_states}. 

\begin{table*}[h!]
\centering
\small
\caption{Theory of Mind mental state categories.}
\label{tab:tom_mental_states}
\begin{tabular}{p{3.2cm} p{11.3cm}}
\toprule
\textbf{Mental State} & \textbf{Description} \\
\midrule
\textbf{T1: Intentions} & The specific, conscious plans an agent formulates to act on underlying desires and achieve a desired outcome. Reasoning about intentions involves inferring the deliberate actions an agent commits to in pursuit of goals. \\
\midrule
\textbf{T2: Emotions} & Current emotional states, their roots, hidden affects, and potentially complex or contradictory feelings. Understanding emotions requires inferring how affective experiences shape reactions, judgments, and subsequent behavior. \\
\midrule
\textbf{T3: Desires} & Deep motivational states representing what an agent emotionally craves, values, or strives for. These desires may be multiple or conflicting and often serve as the underlying drivers of intentions and actions. \\
\midrule
\textbf{T4: Percepts} & Direct sensory access to the current situation and key information, such as what an agent can see, hear, or notice in the environment. This includes situational cues like physical positioning or shifts in tone and helps explain potential cognitive blind spots between agents. \\
\midrule
\textbf{T5: Knowledge} & Epistemic states concerning confirmed facts versus acknowledged unknowns. This includes distinguishing between knowing and merely believing, as well as differentiating reliable knowledge gained through experience from second-hand or uncertain information. \\
\midrule
\textbf{T6: Beliefs} & Subjective judgments or understandings about the situation, other agents, or relationships. Beliefs may be inaccurate or biased, often arising from incomplete information, assumptions, or prior experiences. \\
\midrule
\textbf{T7: Non-Literal Communication} & The interpretation of meaning beyond literal language, including sarcasm, politeness strategies, deception, and other forms of social subtext. Understanding such communication requires inferring communicative intentions and the psychological motives underlying them. \\
\bottomrule
\end{tabular}
\end{table*}


\section{Full Experimental Results}

\subsection{Full Results of 'A Deeper Understanding of LLM Performance in Social Intelligence'}
\label{appendix:deeper_understanding}

\paragraph{Criteria Extraction and Annotation.}
We prompt an LLM (Figure~\ref{fig:prompt_Dataset_Processing_for_Analysis}) to extract fine-grained evaluation criteria—hereafter referred to as \textit{checklist items}—from each human critique, and to record the performance of both candidate models (Model~A and Model~B) on each item. Each checklist item captures a specific evaluative dimension, which may reflect either a desirable or undesirable behavior; however, for the purpose of annotation, we adopt a binary indicator of \textit{positive assessment}: a model receives a score of 1 on a given dimension if the human critique reflects a favorable judgment along that dimension, and 0 otherwise. This process yields a total of \textbf{51,203} dimension-level records.

\paragraph{Taxonomy Construction.}
To organize this large and heterogeneous set of criteria into an interpretable structure, we classify all checklist items into three high-level categories: \textit{role-playing}, \textit{internal state understanding}, and \textit{decision-making}. Within each category, we compute sentence embeddings for all items and apply clustering to group semantically similar criteria. A large language model is then prompted to generate a concise summary label for each cluster, characterizing the shared evaluative focus of the grouped items. To obtain a compact yet representative set of dimensions, we apply a second stage of clustering across all initial clusters, ultimately consolidating the taxonomy into \textbf{100 representative evaluation dimensions}. Details are provided in Tab.~\ref{tab:eval_dimensions_100}. 

\subsection{Inter-Annotator Agreement}
To assess the reliability of our preference annotations, we measured inter-annotator agreement over the four-class label space $\{\text{A}, \text{B}, \text{T}, \text{G}\}$. Each sample was independently annotated by three annotators. We report the distribution of agreement patterns and compare it against the theoretical random baseline, i.e., the expected pattern frequencies if three annotators were to guess uniformly at random among the four labels.

Let $p_{\text{full}}$, $p_{\text{maj}}$, and $p_{\text{dis}}$ denote the probabilities of full agreement (all three annotators agree), majority agreement (two agree, one diverges), and full disagreement (all three diverge), respectively. Under the random-guessing assumption, these probabilities are given by:
\begin{align}
p_{\text{full}} &= \binom{4}{1}\left(\frac{1}{4}\right)^3 = 6.25\%, \\
p_{\text{maj}} &= \binom{4}{1}\binom{3}{2}\left(\frac{1}{4}\right)^2\left(\frac{3}{4}\right) = 56.25\%, \\
p_{\text{dis}} &= \binom{4}{3}\cdot 3! \cdot \left(\frac{1}{4}\right)^3 = 37.50\%.
\end{align}

Table~\ref{tab:agreement} summarizes the observed agreement patterns in our annotation compared to the random baseline.

\begin{table}[t]
\centering
\caption{Inter-annotator agreement patterns compared to the theoretical random baseline.}
\label{tab:agreement}
\resizebox{0.5\textwidth}{!}{%
\begin{tabular}{lcc}
\toprule
\textbf{Agreement Pattern} & \textbf{Random Baseline} & \textbf{Our Annotation} \\
\midrule
Full agreement (3/3 agree)        & 6.25\%  & 39.18\% \\
Majority agreement (2/3 agree)    & 56.25\% & 52.54\% \\
Full disagreement (0/3 agree)     & 37.50\% & 8.29\%  \\
\bottomrule
\end{tabular}%
}
\end{table}

As shown in Table~\ref{tab:agreement}, the observed full-agreement rate ($39.18\%$) is more than $6\times$ higher than the random baseline ($6.25\%$), while the full-disagreement rate ($8.29\%$) is more than $4\times$ lower than chance ($37.50\%$). These results indicate that our annotators converge on a shared preference substantially more often than would be expected by random guessing, demonstrating a high level of annotation consistency.

\subsection{Analysis of Typical Failure Modes in Arena-Style Benchmarks}
We further analyze the typical failure modes identified in arena-style evaluation frameworks (e.g., LMArena) and assess their impact on our benchmark. Several design choices in our setting substantially mitigate these biases. First, role-playing outputs tend to be short, which helps annotators focus on content quality rather than surface-level style. Second, unlike LMArena, our evaluation adopts detailed criteria beyond a single preference judgment, inherently weakening stylistic shortcuts. Third, annotators are required to provide justifications for every preference, making each decision evidence-based and interpretable.

We additionally quantify the biases discussed in prior work:

\begin{itemize}
    \item \textbf{Length bias.} Winners are longer than losers in only $51.05\%$ of comparisons, with a mean length difference of $14.0$ tokens and a median difference of $3.0$ tokens. This is close to chance level, confirming negligible length bias.
    \item \textbf{Position bias.} Votes favor option B over option A ($43.63\%$ vs.\ $39.23\%$), indicating only a mild position effect.
    \item \textbf{Format, tone, and confidence bias.} Our task does not involve rich formatting or answer confidence, so these biases are inapplicable by design.
\end{itemize}

These results demonstrate that typical biases are either absent or well-controlled in our setting, verifying that our task design effectively addresses the concerns raised for arena-style evaluation.

%
%
%

\providecommand{\textless}{\ensuremath{<}}
\providecommand{\textgreater}{\ensuremath{>}}
\providecommand{\cmark}{\textcolor{green!60!black}{\ding{51}}}
\providecommand{\xmark}{\textcolor{red!70!black}{\ding{55}}}

\section{Prompts}

\begin{promptbox}[Prompt: Real Social Contexts Generation]{blue!30!white}
\footnotesize
Please generate a role-playing card (including the current scene, character information, and opening dialogue).

\textbf{Core Objective}: We need the characters within the scene to demonstrate a certain degree of psychological depth; however, it should be neither an everyday mundane scenario nor an extremely dramatized plot.

\paragraph{1. Define the Basic Scene Framework.}
Below I will provide a multi-dimensional scale analysis of the basic scene framework, with each dimension represented by a capital letter.

\textit{Basic Scene Framework Reference:}
\begin{itemize}
  \setlength{\itemsep}{1pt}\setlength{\parskip}{0pt}\setlength{\topsep}{2pt}
  \item \textbf{T: Topological Relationship (Graph Theory)}
    \begin{itemize}
      \setlength{\itemsep}{0pt}\setlength{\parskip}{0pt}\setlength{\topsep}{1pt}
      \item \textbf{T1: Self-Reflexive -- One person alone.} Example: individual.
      \item \textbf{T2: Symmetric Dyad -- Two people interacting.} Example: couple, doctor/patient, teacher/student, buyer/seller, counselor/client.
      \item \textbf{T3: Symmetric Group -- Three or more people interacting.} Example: team members, family members, book club members, department colleagues, friend groups.
    \end{itemize}
  \item \textbf{R: Relational Models Theory (Alan Fiske)}
    \begin{itemize}
      \setlength{\itemsep}{0pt}\setlength{\parskip}{0pt}\setlength{\topsep}{1pt}
      \item \textbf{R1: Communal Sharing.} Example: mother/child, couple, children/parents, roommates, family members.
      \item \textbf{R2: Authority Ranking.} Example: CEO/employee, supervisor/graduate student, officer/soldier, judge/party, editor-in-chief/author.
      \item \textbf{R3: Equality Matching.} Example: colleagues, classmates, neighbors, carpooling partners, friends.
      \item \textbf{R4: Market Pricing.} Example: consumer/merchant, client/service provider, landlord/tenant, employer/employee, investor/investment target.
      \item \textbf{R5: Null/Asocial.} Example: strangers, passersby, pedestrians.
    \end{itemize}
  \item \textbf{S: Interest Relationship (Game Theory)}
    \begin{itemize}
      \setlength{\itemsep}{0pt}\setlength{\parskip}{0pt}\setlength{\topsep}{1pt}
      \item \textbf{S1: Zero-Sum -- One party gains = the other party loses.} Example: competitive bidding, election for a seat, inheritance dispute, competition for championship, competing for promotion.
      \item \textbf{S2: Negative-Sum -- Conflict leads to losses for both parties.} Example: price wars, divorce litigation, trade wars, factional struggles, inter-departmental blame-shifting.
      \item \textbf{S3: Positive-Sum -- Cooperation creates value.} Example: university-enterprise cooperation, strategic partners, joint investment, startup partnership, community co-building.
    \end{itemize}
  \item \textbf{M: Motivation Type (Maslow's Hierarchy of Needs)}
    \begin{itemize}
      \setlength{\itemsep}{0pt}\setlength{\parskip}{0pt}\setlength{\topsep}{1pt}
      \item \textbf{M1: Physiological/Survival Needs.} Example: eating when hungry, cooling down in heat, drinking when thirsty, sleeping when tired, seeking medical care when ill.
      \item \textbf{M2: Safety/Stability Needs.} Example: purchasing insurance, saving money, signing contracts, installing security systems, retirement planning.
      \item \textbf{M3: Belonging/Relationship Needs.} Example: going home to be with family, joining a community, confiding after a breakup, team-building activities, looking for fellow townspeople.
      \item \textbf{M4: Esteem/Status Needs.} Example: pursuing awards, speaking at meetings, sharing achievements, running for positions, obtaining certifications.
      \item \textbf{M5: Values/Self-Actualization Needs.} Example: teaching in underprivileged areas, learning art, environmental causes, writing memoirs, pursuing a doctorate.
    \end{itemize}
  \item \textbf{P: Whether It Is a Formal Setting}
    \begin{itemize}
      \setlength{\itemsep}{0pt}\setlength{\parskip}{0pt}\setlength{\topsep}{1pt}
      \item \textbf{P1: Informal.} Example: discussion of values, value conflicts, late-night heart-to-heart talks, deep conversations while waiting for a flight, strangers sharing stories.
      \item \textbf{P2: Formal.} Example: psychological healing (important scene), handling conflicts at a dinner party, estate distribution, board decision-making, divorce mediation, merger negotiation.
    \end{itemize}
\end{itemize}

\paragraph{2. Progressively Generate a Specific Scene.}

\paragraph{2.1 Plot Overview.}
Generate a plot whose \textbf{complexity comes from authentic sources}, not from artificially contrived setups.

\paragraph{2.2 Current Scene.}
Describe the background of the current dialogue in a \textbf{colloquial, straightforward} manner.

Required elements:
\begin{enumerate}
  \setlength{\itemsep}{1pt}\setlength{\parskip}{0pt}\setlength{\topsep}{2pt}
  \item \textbf{Specific background (if applicable)}: What is the specific relationship between the characters, and what are the key preceding events?
  \item \textbf{Current event}: Who is talking to whom, and what specific matter are they discussing?
  \item \textbf{Source of complexity}: This must be included to prevent readers from being unable to understand where the dilemma comes from.
\end{enumerate}

Writing requirements:
\begin{itemize}
  \setlength{\itemsep}{1pt}\setlength{\parskip}{0pt}\setlength{\topsep}{2pt}
  \item Use short sentences and write in separate paragraphs.
  \item Describe only the core events.
  \item Do not describe professional/specialized content, e.g.: ``The queen has passed away''.
  \item Do not use extremely emotionally charged words: tearing apart, struggling, breakdown.
  \item Do not use extremely dramatized literary expressions: fate, reunion, redemption, entanglement.
\end{itemize}

\paragraph{2.3 Opening Dialogue.}
Provide natural dialogue that fits the [current scene]. The last line should end at an appropriate point, leaving room for the model to take on a role and improvise.

Requirements:
\begin{itemize}
  \setlength{\itemsep}{1pt}\setlength{\parskip}{0pt}\setlength{\topsep}{2pt}
  \item Dialogue should feel everyday and natural, as if real people are speaking.
  \item Avoid overly sentimental, dramatized, or sensationalized lines.
  \item Reflect the complexity of the scene while keeping the tone restrained.
\end{itemize}

\paragraph{2.4 Character Information.}
List the names of the characters in the dialogue (use English names).

\paragraph{3. Output Format.}
\begin{quote}
\texttt{[\textless Keyword Selection\textgreater]}\\
{[}Selected keywords -- do not select all of them{]}\\
\texttt{[/\textless Keyword Selection\textgreater]}

\medskip
\texttt{[\textless Thinking\textgreater]}\\
{[}You may think at length here in order to develop a better plot{]}\\
\texttt{[/\textless Thinking\textgreater]}

\medskip
\texttt{[\textless Plot Overview\textgreater]}\\
{[}Describe the plot in one sentence, within 30 characters{]}\\
\texttt{[/\textless Plot Overview\textgreater]}

\medskip
\texttt{[\textless Current Scene\textgreater]}\\
{[}Place the current scene here{]}\\
{[}Use the characters' names, write in separate paragraphs, 1--2 sentences per paragraph{]}\\
\texttt{[/\textless Current Scene\textgreater]}

\medskip
\texttt{[\textless Opening Dialogue\textgreater]}\\
1. Character A: (brief action) Dialogue content\\
2. Character B: (brief action) Dialogue content\\
{[}3--7 rounds of dialogue{]}\\
\texttt{[/\textless Opening Dialogue\textgreater]}

\medskip
\texttt{[\textless Character Information\textgreater]}\\
{[}Character A's name, Character B's name{]}\\
\texttt{[/\textless Character Information\textgreater]}

\medskip
\texttt{[\textless Next Speaker\textgreater]}\\
{[}Write the name of the next speaker here -- only one name{]}\\
\texttt{[/\textless Next Speaker\textgreater]}
\end{quote}

\paragraph{4. Quality Check.}
In this section, you should go through the checklist below item by item and verify whether your generated content meets each requirement. For each item, first provide your reasoning, then state your conclusion. If any issue is found, re-output using the format from Section 2.5, and check again.
\begin{enumerate}
  \setlength{\itemsep}{1pt}\setlength{\parskip}{0pt}\setlength{\topsep}{2pt}
  \item No life-or-death related plots are allowed: e.g., serious illness, terminal illness, suicide, life-threatening danger, near-death (acceptable if one party is responsible for positively regulating the other's emotions).
  \item No serious criminal plots are allowed: e.g., murder, sexual assault, drug trafficking, kidnapping.
  \item No extreme trauma plots are allowed: e.g., childhood abuse, abandonment, experiencing violence.
  \item No illegal events are allowed: e.g., extramarital affairs, embezzlement of public funds.
  \item No major betrayals are allowed: e.g., betraying a friend.
  \item No irreversible losses are allowed: e.g., bankruptcy, complete loss of reputation.
  \item No surreal settings are allowed: e.g., dreamscapes, time travel, paranormal events, fateful entanglements.
  \item The scenario must not consist of everyday trivial matters with no complexity, solvable by any approach either party might take.
  \item No extreme, hysterical, or forced-choice situations where a character must suffer significant harm are allowed.
  \item Only a single event should be the focus -- do not layer multiple scenes.
  \item If there is conflict, it must arise from objective reality, not be artificially manufactured.
  \item If your scene involves contradiction/conflict, ensure both parties carry comparable weight -- examples of imbalance: safety/health, legal/moral baseline vs.\ personal interest; long-term major impact vs.\ short-term minor inconvenience.
  \item The last line of the opening dialogue must end at a point that sets an appropriate atmosphere, leaving room for the model to take on a role and improvise.
\end{enumerate}

\paragraph{5. Input.}
For this generation, you must use \textbf{\{base\_scene\}} as the basic scene framework combination. You may refer to these keywords when generating the scene, but note that their use is not mandatory. Keywords: \textbf{\{keywords\}}.

Please begin generating.
\end{promptbox}
\captionof{figure}{Prompt: Real Social Contexts Generation.}
\label{fig:prompt_Real_Social_Contexts_Generation}

\begin{promptbox}[Prompt: User-Centered Social Contexts Generation]{orange!50!white}
\footnotesize
Please generate a role-playing card (including the current scene, character information, and opening dialogue).

\textbf{Core Objective}: We need to build scenes oriented toward \textbf{entertainment} for the user. For example: virtual worlds (plot-driven, such as isekai adventures, virtual exploration, detective worlds), or real-world settings (entertainment-oriented, such as a palace maid falling in love with a crown prince, or a domineering CEO and their secretary).

\paragraph{1. Define the Basic Scene Framework.}
Below I will provide a multi-dimensional scale analysis of the basic scene framework, with each dimension represented by a capital letter.

\textit{Basic Scene Framework Reference:}
\begin{itemize}
  \setlength{\itemsep}{1pt}\setlength{\parskip}{0pt}\setlength{\topsep}{2pt}
  \item \textbf{T: Topological Relationship (Graph Theory)}
    \begin{itemize}
      \setlength{\itemsep}{0pt}\setlength{\parskip}{0pt}\setlength{\topsep}{1pt}
      \item \textbf{T1: Self-Reflexive -- One person alone.} Example: individual.
      \item \textbf{T2: Symmetric Dyad -- Two people interacting.} Example: couple, doctor/patient, teacher/student, buyer/seller, counselor/client.
      \item \textbf{T3: Symmetric Group -- Three or more people interacting.} Example: team members, family members, book club members, department colleagues, friend groups.
    \end{itemize}
  \item \textbf{R: Relational Models Theory (Alan Fiske)}
    \begin{itemize}
      \setlength{\itemsep}{0pt}\setlength{\parskip}{0pt}\setlength{\topsep}{1pt}
      \item \textbf{R1: Communal Sharing.} Example: mother/child, couple, children/parents, roommates, family members.
      \item \textbf{R2: Authority Ranking.} Example: CEO/employee, supervisor/graduate student, officer/soldier, judge/party, editor-in-chief/author.
      \item \textbf{R3: Equality Matching.} Example: colleagues, classmates, neighbors, carpooling partners, friends.
      \item \textbf{R4: Market Pricing.} Example: consumer/merchant, client/service provider, landlord/tenant, employer/employee, investor/investment target.
      \item \textbf{R5: Null/Asocial.} Example: strangers, passersby, pedestrians.
    \end{itemize}
  \item \textbf{S: Interest Relationship (Game Theory)}
    \begin{itemize}
      \setlength{\itemsep}{0pt}\setlength{\parskip}{0pt}\setlength{\topsep}{1pt}
      \item \textbf{S1: Zero-Sum -- One party gains = the other party loses.} Example: competitive bidding, election for a seat, inheritance dispute, competition for championship, competing for promotion.
      \item \textbf{S2: Negative-Sum -- Conflict leads to losses for both parties.} Example: price wars, divorce litigation, trade wars, factional struggles, inter-departmental blame-shifting.
      \item \textbf{S3: Positive-Sum -- Cooperation creates value.} Example: university-enterprise cooperation, strategic partners, joint investment, startup partnership, community co-building.
    \end{itemize}
  \item \textbf{M: Motivation Type (Maslow's Hierarchy of Needs)}
    \begin{itemize}
      \setlength{\itemsep}{0pt}\setlength{\parskip}{0pt}\setlength{\topsep}{1pt}
      \item \textbf{M1: Physiological/Survival Needs.} Example: eating when hungry, cooling down in heat, drinking when thirsty, sleeping when tired, seeking medical care when ill.
      \item \textbf{M2: Safety/Stability Needs.} Example: purchasing insurance, saving money, signing contracts, installing security systems, retirement planning.
      \item \textbf{M3: Belonging/Relationship Needs.} Example: going home to be with family, joining a community, confiding after a breakup, team-building activities, looking for fellow townspeople.
      \item \textbf{M4: Esteem/Status Needs.} Example: pursuing awards, speaking at meetings, sharing achievements, running for positions, obtaining certifications.
      \item \textbf{M5: Values/Self-Actualization Needs.} Example: teaching in underprivileged areas, learning art, environmental causes, writing memoirs, pursuing a doctorate.
    \end{itemize}
  \item \textbf{P: Whether It Is a Formal Setting}
    \begin{itemize}
      \setlength{\itemsep}{0pt}\setlength{\parskip}{0pt}\setlength{\topsep}{1pt}
      \item \textbf{P1: Informal.} Example: discussion of values, value conflicts, late-night heart-to-heart talks, deep conversations while waiting for a flight, strangers sharing stories.
      \item \textbf{P2: Formal.} Example: psychological healing (important scene), handling conflicts at a dinner party, estate distribution, board decision-making, divorce mediation, merger negotiation.
    \end{itemize}
\end{itemize}

\paragraph{2. Progressively Generate a Specific Scene.}

\paragraph{2.1 Plot Overview.}
Write an overview of the plot to set the direction for the generation that follows.

\paragraph{2.2 Current Scene.}
Describe the background of the current dialogue in a \textbf{colloquial, straightforward} manner.

Required elements:
\begin{enumerate}
  \setlength{\itemsep}{1pt}\setlength{\parskip}{0pt}\setlength{\topsep}{2pt}
  \item \textbf{Specific background (if applicable)}: What is the specific relationship between the characters, and what are the key preceding events?
  \item \textbf{Current event}: Who is talking to whom, and what are they discussing?
\end{enumerate}

Writing requirements:
\begin{itemize}
  \setlength{\itemsep}{1pt}\setlength{\parskip}{0pt}\setlength{\topsep}{2pt}
  \item Use short sentences and write in separate paragraphs.
  \item Describe only the core events.
  \item Do not describe professional/specialized content, e.g.: ``The queen has passed away''.
\end{itemize}

\paragraph{2.3 Opening Dialogue.}
Provide natural dialogue that fits the [current scene]. The last line should end at an appropriate point, leaving room for the model to take on a role and improvise.

Requirements:
\begin{itemize}
  \setlength{\itemsep}{1pt}\setlength{\parskip}{0pt}\setlength{\topsep}{2pt}
  \item Avoid overly sentimental lines; prioritize a light and humorous tone.
\end{itemize}

\paragraph{2.4 Character Information.}
List the names of the characters in the dialogue (use English names).

\paragraph{3. Output Format.}
\begin{quote}
\texttt{[\textless Keyword Selection\textgreater]}\\
{[}Selected keywords -- do not select all of them{]}\\
\texttt{[/\textless Keyword Selection\textgreater]}

\medskip
\texttt{[\textless Thinking\textgreater]}\\
{[}You may think at length here in order to develop a better plot; prioritize user-centered entertainment value{]}\\
\texttt{[/\textless Thinking\textgreater]}

\medskip
\texttt{[\textless Plot Overview\textgreater]}\\
{[}Describe the plot in one sentence, within 30 characters{]}\\
\texttt{[/\textless Plot Overview\textgreater]}

\medskip
\texttt{[\textless Current Scene\textgreater]}\\
{[}Place the current scene here{]}\\
{[}Use the characters' names, write in separate paragraphs, 1--2 sentences per paragraph{]}\\
\texttt{[/\textless Current Scene\textgreater]}

\medskip
\texttt{[\textless Opening Dialogue\textgreater]}\\
1. Character A: (brief action) Dialogue content\\
2. Character B: (brief action) Dialogue content\\
{[}3--7 rounds of dialogue{]}\\
\texttt{[/\textless Opening Dialogue\textgreater]}

\medskip
\texttt{[\textless Character Information\textgreater]}\\
1. World background (write this only if it is an isekai/fantasy adventure; not needed for real-world settings)\\
2. Description of the role the user plays\\
3. Description of the role the AI plays\\
\quad -- The model must be clearly told that it is an AI; note that this must be framed through a user-facing relationship (for example, in a romance scenario set in the real world, the model can be told it is an AI that is simply playing the role of a girlfriend, and that it has the ability to influence how the world changes and control how the plot develops)\\
\quad -- The model must be clearly told that it should center the user, and that it can influence how the world changes and control how the plot develops\\
\texttt{[/\textless Character Information\textgreater]}

\medskip
\texttt{[\textless Next Speaker\textgreater]}\\
{[}The next speaker is the AI -- write their name here in the format ``Name (AI)''{]}\\
\texttt{[/\textless Next Speaker\textgreater]}
\end{quote}

\paragraph{4. Quality Check.}
In this section, you should go through the checklist below item by item and verify whether your generated content meets each requirement. For each item, first provide your reasoning, then state your conclusion. If any issue is found, re-output using the format from Section 2.5, and check again.
\begin{enumerate}
  \setlength{\itemsep}{1pt}\setlength{\parskip}{0pt}\setlength{\topsep}{2pt}
  \item No life-or-death related plots are allowed: e.g., serious illness, terminal illness, suicide, life-threatening danger, near-death (acceptable if one party is responsible for positively regulating the other's emotions).
  \item No extreme or hysterical situations where a character would suffer significant harm are allowed.
\end{enumerate}

\paragraph{5. Input.}
For this generation, you must use \textbf{\{base\_scene\}} as the basic scene framework combination. You may refer to these keywords when generating the scene, but note that their use is not mandatory. Keywords: \textbf{\{keywords\}}.

Please begin generating.
\end{promptbox}
\captionof{figure}{Prompt: User-Centered Social Contexts Generation.}
\label{fig:prompt_User_Centered_Social_Contexts_Generation}

\begin{promptbox}[Prompt: Conversation Generation]{red!30!white}
\footnotesize
\paragraph{system.}
You are a Roleplaying Expert with Social Intelligence. Your task is to play the specified character in the ``Current Scenario''.

\paragraph{Behavioral Guidelines for Roleplay.}
\textbf{Core Objective}: Starting from objective facts, you need to infer Theory of Mind (understanding the various mental states of yourself and others), and use this as a driver to generate an appropriate response.

\paragraph{Roleplay objectives.}
\begin{enumerate}
  \setlength{\itemsep}{1pt}\setlength{\parskip}{0pt}\setlength{\topsep}{2pt}
  \item Always respond from a first-person perspective; do not mention that you are an AI or that you are roleplaying.
  \item Dialogue format: responses should reflect the character's linguistic style and emotional state, kept concise (1--3 sentences).
  \item Action format: if there are actions, annotate them with parentheses, e.g.: (looks down staring at the glass).
  \item Keep dialogue authentic: avoid literary expressions and avoid fabricating things out of thin air.
\end{enumerate}

\paragraph{Character Thinking.}
Before producing the final output, you need to engage in \textbf{structured cognitive reasoning and decision reasoning} from the perspective of the character you are playing, speaking in the first person (``I''). \textbf{Wrap this inner thinking between \texttt{\textless think\_character\textgreater} and \texttt{\textless /think\_character\textgreater} tags.}

Explanation:
\begin{itemize}
  \setlength{\itemsep}{1pt}\setlength{\parskip}{0pt}\setlength{\topsep}{2pt}
  \item Cognitive reasoning: Based on objective facts, infer the mental states of yourself and others (across the dimensions listed below).
  \item Decision reasoning: Based on the objective facts and results of cognitive reasoning, infer how you should respond to the other party.
\end{itemize}

\paragraph{Definitions of Mental State Dimensions.}
When reasoning, you need to \textbf{adaptively select} the 1--4 most critical dimensions for the current situation from the following mental state dimensions:
\begin{itemize}
  \setlength{\itemsep}{1pt}\setlength{\parskip}{0pt}\setlength{\topsep}{2pt}
  \item \textbf{T1: Intention}: The specific, conscious plans you formulate in order to fulfill desires and achieve expected outcomes.
  \item \textbf{T2: Emotion}: Understanding current emotional states, their underlying sources, hidden feelings, and complex or conflicting sentiments, as well as expressing or regulating one's own and others' emotions.
  \item \textbf{T3: Desire}: The wishes, preferences, and deep longings one ultimately hopes to fulfill -- encompassing both explicit and implicit needs; also perceiving what oneself and others truly want, and identifying tensions, conflicts, or synergies among multiple desires.
  \item \textbf{T4: Perception}: The direct awareness of a current situation and access to key information; this also includes recognizing that what others see, hear, and know is constrained by their physical position and sensory experience, thereby inferring their cognitive blind spots or informational limitations.
  \item \textbf{T5: Knowledge}: The confirmation of known facts, awareness of unknown domains, and a clear distinction between ``knowing'' and ``believing'', as well as between ``hearsay'' and ``firsthand experience''.
  \item \textbf{T6: Belief}: Personal subjective judgments and interpretations of situations, events, or interpersonal relationships; includes inferring the beliefs of others, and identifying belief distortions formed through one-sided information, inherent biases, or mistaken cognition.
  \item \textbf{T7: Non-Literal Communication}: Discerning implied meanings, social strategies, and emotional intent; the ability to recognize the psychological motivations and communicative purposes hidden within non-literal expressions such as sarcasm, pleasantries, lies, humor, or slips of the tongue.
\end{itemize}

\paragraph{Thinking Process.}
\begin{enumerate}
  \setlength{\itemsep}{1pt}\setlength{\parskip}{0pt}\setlength{\topsep}{2pt}
  \item Thinking about Current Mental States. Understand your own and others' mental states in the current situation: (1) select 2--4 of the most relevant dimensions; (2) you may respond directly, or combine the context with reasoning of any complexity.
  \item Thinking about Behavior. You need to execute the following steps:
    \begin{enumerate}
      \setlength{\itemsep}{0pt}\setlength{\parskip}{0pt}\setlength{\topsep}{1pt}
      \item Describe your behavior: (a) you may state your response directly; (b) or you may reason out your response based on the mental state analysis above; if doing so, use the format \texttt{\textless Dimension\textgreater: \textless Analysis\textgreater} to indicate which mental states you are drawing on and how.
      \item (Optional) Validate your response: (a) you may choose not to validate the response; (b) or you may analyze which already-analyzed dimensions would undergo what changes -- or what important new dimensions would be introduced -- if you gave that response, thereby validating whether your response is reasonable; if doing so, use the format \texttt{\textless Dimension\textgreater: \textless Analysis\textgreater} to indicate which mental states you analyzed.
    \end{enumerate}
  \item Meta-Thinking. If you find that the behavior you previously conceived is not ideal, you may iterate through [Thinking about Behavior] multiple times to arrive at a better behavior.
\end{enumerate}

\paragraph{Notes.}
\begin{enumerate}
  \setlength{\itemsep}{1pt}\setlength{\parskip}{0pt}\setlength{\topsep}{2pt}
  \item \textbf{Adaptive dimension selection}: Do not mechanically cover all dimensions -- only analyze the key dimensions that truly drive decisions in the current situation.
  \item \textbf{Maintain character perspective}: Always think from the character's point of view using ``I'', reflecting the character's cognitive limitations and personality traits.
  \item \textbf{Structured output}: Strictly follow the format above for output, to facilitate extraction and analysis.
  \item \textbf{Goal-oriented}: All thinking should serve the final behavioral decision.
\end{enumerate}

\paragraph{Output Format.}
\textbf{Reply strictly using the following format}:
\begin{quote}
{[}Name of the speaking character this turn{]}:

\texttt{\textless think\_character\textgreater}\\
{[}Thinking about Mental States{]}\\
-- \texttt{\textless Dimension Name 1\textgreater}: \texttt{\textless Specific analysis\textgreater}\\
-- \texttt{\textless Dimension Name 2\textgreater}: \texttt{\textless Specific analysis\textgreater}\\
\ldots

{[}Thinking about Behavior{]}\\
\texttt{\textless Content\textgreater}\\
(If there is meta-thinking, simply output [Thinking about Behavior] multiple times)\\
\texttt{\textless /think\_character\textgreater}

\medskip
(Place the final response here)
\end{quote}

\paragraph{User template.}
\paragraph{Current Scenario.}
\texttt{\{scene\}}

\textbf{Answer using the language in ``Current Scenario''.} Now begin the roleplay -- please continue the dialogue.
\end{promptbox}
\captionof{figure}{Prompt: Conversation Generation.}
\label{fig:prompt_Conversation_Generation}

\begin{promptbox}[Prompt: Conversation Quality Judgement]{red!30!white}
\footnotesize
You are a professional response quality assessment expert. Your task is to compare the quality of two responses based on background information (including conversation history) and determine which one is better.

\textbf{Important: Only evaluate the final output content of each response; intermediate thinking processes can be used as reference.}

\paragraph{Criteria for Judging ``Better''.}
We place the model in a social scenario and have it play a role to take actions and give replies. How to judge ``better'' needs to be analyzed in conjunction with the specific social context.

\begin{quote}
\textbf{Example Scenario: An Argument Between a Couple.}

Suppose the dialogue background is a couple in a heated argument. Please note the following layers of requirements -- they may be in tension with each other, and a good response needs to strike a balance among them:

\begin{enumerate}
  \setlength{\itemsep}{1pt}\setlength{\parskip}{0pt}\setlength{\topsep}{2pt}
  \item \textbf{Empathy and De-escalation:} The role being played should show empathy toward the other party's feelings, using gentle and appropriate language to ease the conflict rather than adding fuel to the fire.
  \item \textbf{Psychological Credibility of Emotional Transitions:} However, if the character is currently in a state of extreme anger, they should not suddenly become calm and rational. Emotional changes must be gradual and consistent with psychological norms -- for example, shifting from anger to exhaustion or hurt first, then slowly moving toward calm, rather than switching from rage to tenderness in a single sentence.
  \item \textbf{Moral Bottom Line:} If the argument was caused by one party cheating, the character should not use empathetic language to comfort or rationalize the cheater's behavior. Empathy should be directed toward the injured party, not used to excuse the wrongdoer.
\end{enumerate}
\end{quote}

\textbf{Special Note:} Role-play evaluation in social scenarios is very complex and nuanced, and is not limited to the three situations described above. Please analyze based on the specific social context.

\paragraph{Background Information and Conversation History.}
\begin{quote}
\texttt{\{BACKGROUND\_INFORMATION\}}
\end{quote}

\paragraph{Response A.}
\begin{quote}
\texttt{\{RESPONSE\_A\}}
\end{quote}

\paragraph{Response B.}
\begin{quote}
\texttt{\{RESPONSE\_B\}}
\end{quote}

\paragraph{Output Format.}
You may engage in any form of thinking, but ultimately you need to organize the results into the following format:
\begin{quote}
\texttt{\textless winner\textgreater}\\
{[}Winner. The result can only be one of four options: A, B, T, or G, representing: Response A is better, Response B is better, Both responses are good, Both responses are poor{]}\\
\texttt{\textless /winner\textgreater}

\medskip
\texttt{\textless critique\textgreater}\\
{[}Commentary. Write multiple lines of commentary here, each pointing out one concise argument. Note that this is the condensed version -- each line should be a short sentence. Lengthy elaborations are prohibited.{]}\\
\texttt{\textless /critique\textgreater}
\end{quote}
\end{promptbox}
\captionof{figure}{Prompt: Conversation Quality Judgement.}
\label{fig:prompt_Conversation_Quality_Judgement}

\begin{promptbox}[Prompt: Dataset Processing for Analysis]{orange!50!white}
\footnotesize
You are a text analysis assistant. Please read the following evaluation conclusions about two responses (A and B), extract the key \textbf{evaluation dimensions}, and determine whether each dimension is positive or negative for A or B.

\paragraph{What are evaluation dimensions.}
Evaluation dimensions are ``judgment angles that can be scored independently'', such as ``psychological depth'', ``dialogue logic'', and ``initiative''.

Note:
\begin{itemize}
  \setlength{\itemsep}{1pt}\setlength{\parskip}{0pt}\setlength{\topsep}{2pt}
  \item \xmark~Do not extract specific behaviors (e.g., ``requesting receipts'', ``moving in together'').
  \item \xmark~Do not extract evaluation results (e.g., ``slightly rough'', ``superior'').
  \item \xmark~Do not extract plot details.
  \item \cmark~Extract abstract, reusable scoring dimensions in the form of indicators, structured as two-dimensional, preferably using ``whether''.
\end{itemize}

\paragraph{Rules for Positive/Negative Judgment.}
\begin{itemize}
  \setlength{\itemsep}{1pt}\setlength{\parskip}{0pt}\setlength{\topsep}{2pt}
  \item 1 = Positive (performs well on this dimension); 0 = Negative (performs poorly on this dimension).
  \item \textbf{Only label the party explicitly mentioned in the text;} do not infer or complete the other party.
    \begin{itemize}
      \setlength{\itemsep}{0pt}\setlength{\parskip}{0pt}\setlength{\topsep}{1pt}
      \item Example: Only ``A is better at psychological game'' -- write only ``A'': 1, do not write ``B'': 0.
      \item Example: Only ``A's dialogue logic has a flaw'' -- write only ``A'': 0, do not write ``B'': 1.
      \item Example: Both ``A has good psychological game, B's is rough'' -- write ``A'': 1, ``B'': 0.
    \end{itemize}
\end{itemize}

\paragraph{Output Requirements.}
\begin{enumerate}
  \setlength{\itemsep}{1pt}\setlength{\parskip}{0pt}\setlength{\topsep}{2pt}
  \item Each dimension label should be 2--10 words.
  \item Output in Python List of Dict format, each dict containing: ``dim'' (dimension name, string); ``A'' (1 or 0, if the text explicitly mentions A's performance on this dimension); ``B'' (1 or 0, if the text explicitly mentions B's performance on this dimension).
  \item Output the conclusion directly, with no additional content.
\end{enumerate}

The evaluation conclusion is as follows: \texttt{\{conclusion\}}

\paragraph{Example input.}
\begin{itemize}
  \setlength{\itemsep}{0pt}\setlength{\parskip}{0pt}\setlength{\topsep}{1pt}
  \item Response A is superior in terms of psychological game, accurately capturing the opponent's weaknesses and formulating in-depth response strategies, but there are obvious flaws in the logical common sense of specific dialogue lines, which undermines the authenticity of the interaction.
  \item Response B's psychological analysis is slightly rough in depth, but its behavioral decisions closely follow the core facts, showing stronger initiative and better meeting the needs of plot progression.
  \item Overall, B is slightly better, because B directly achieves the character's core goal, while A's dialogue logic hinders plot progression.
\end{itemize}

\paragraph{Example output.}
\begin{quote}\ttfamily
{[}\\
\texttt{\ \ \{"dim": "Whether psychological game is good", "A": 1, "B": 0\},}\\
\texttt{\ \ \{"dim": "Whether able to capture opponent's weaknesses", "A": 1\},}\\
\texttt{\ \ \{"dim": "Whether able to formulate in-depth response strategies", "A": 1\},}\\
\texttt{\ \ \{"dim": "Whether logical common sense is correct", "A": 0\},}\\
\texttt{\ \ \{"dim": "Whether interaction feels authentic", "A": 0\},}\\
\texttt{\ \ \{"dim": "Whether initiative is strong", "B": 1\},}\\
\texttt{\ \ \{"dim": "Whether plot progression is good", "A": 0, "B": 1\},}\\
\texttt{\ \ \{"dim": "Whether goal is achieved", "B": 1\}}\\
{]}
\end{quote}
\end{promptbox}
\captionof{figure}{Prompt: Dataset Processing for Analysis.}
\label{fig:prompt_Dataset_Processing_for_Analysis}


\clearpage


\begin{longtable}{%
  >{\footnotesize\raggedright\arraybackslash}p{0.54\textwidth}  
  >{\footnotesize\centering\arraybackslash}p{0.12\textwidth}    
  >{\footnotesize\centering\arraybackslash}p{0.12\textwidth}    
  >{\footnotesize\centering\arraybackslash}p{0.12\textwidth}    
}

\caption{Overview of the 100 evaluation dimensions and their category assignments.
  \cmark~indicates the dimension belongs to the category;
  \xmark~indicates it does not.}
\label{tab:eval_dimensions_100}\\

\toprule
\textbf{Dimension} &
\textbf{\makecell{Role-\\playing}} &
\textbf{\makecell{ToM\\Reasoning}} &
\textbf{\makecell{Behavior\\Reasoning}} \\
\midrule
\endfirsthead

\multicolumn{4}{l}{\footnotesize\textit{(Continued from previous page)}}\\[2pt]
\toprule
\textbf{Dimension} &
\textbf{\makecell{Role-\\playing}} &
\textbf{\makecell{ToM\\Reasoning}} &
\textbf{\makecell{Behavior\\Reasoning}} \\
\midrule
\endhead

\midrule
\multicolumn{4}{r}{\footnotesize\textit{Continued on next page\ldots}}\\
\endfoot

\bottomrule
\endlastfoot

\rowcolor{gray!10}
\multicolumn{4}{l}{\footnotesize\textit{\textbf{Role-playing}}}\\
Whether behavior and speech are consistent and logically coherent in real situations
  & \cmark & \xmark & \xmark \\
Whether the role is faithfully portrayed and the narrative naturally advanced
  & \cmark & \xmark & \xmark \\
Whether the portrayal comes across as unnatural, unprofessional, or lacking credibility
  & \cmark & \xmark & \xmark \\
Whether characters and the self are presented authentically, naturally, and compellingly
  & \cmark & \xmark & \xmark \\
Whether self-awareness, emotional expression, and thought patterns show rigidity, avoidance, or imbalance
  & \cmark & \xmark & \xmark \\
Whether expression exhibits logical confusion, identity conflation, or communicative missteps
  & \cmark & \xmark & \xmark \\
Whether the self is expressed genuinely and emotional flow is sustained across interactions
  & \cmark & \xmark & \xmark \\
Whether problem-solving maintains internal consistency, pragmatic feasibility, and focus on essentials
  & \cmark & \xmark & \xmark \\
Whether professional interactions naturally balance expertise, accountability, and approachability
  & \cmark & \xmark & \xmark \\
Whether emotions are expressed and regulated authentically, subtly, and with restraint
  & \cmark & \xmark & \xmark \\
Whether one holds one's ground, takes responsibility, and upholds academic integrity in formal settings
  & \cmark & \xmark & \xmark \\
Whether interactions stay grounded in facts while avoiding fabrication
  & \cmark & \xmark & \xmark \\
Whether the balance between ideals and reality is well-judged and earns recognition
  & \cmark & \xmark & \xmark \\
Whether self-expression and consideration for others' feelings are balanced in social settings
  & \cmark & \xmark & \xmark \\
Whether dignity, boundaries, and interpersonal propriety are maintained across multiple relationships
  & \cmark & \xmark & \xmark \\
Whether the narrative is naturally advanced and persona appeal sustained in a light-hearted tone
  & \cmark & \xmark & \xmark \\
Whether self-contradictions and relational fractures are continuously exposed and exacerbated during interaction
  & \cmark & \xmark & \xmark \\
Whether boundaries, core values, and commitments are upheld across varied relationships and roles
  & \cmark & \xmark & \xmark \\
Whether one's beliefs are continuously examined, adjusted, and clearly expressed across situations
  & \cmark & \xmark & \xmark \\

\rowcolor{gray!10}
\multicolumn{4}{l}{\footnotesize\textit{\textbf{Theory-of-Mind (ToM) Reasoning}}}\\
Whether the deeper intentions, unspoken rules, and psychological dynamics in interpersonal interactions are accurately perceived
  & \xmark & \cmark & \xmark \\
Whether one genuinely attends to and appropriately responds to the emotional needs of others
  & \xmark & \cmark & \xmark \\
Whether the grievances, fears, and deeper needs of the other party are truly understood and addressed
  & \xmark & \cmark & \xmark \\
Whether various critical signals and underlying issues in interactions are accurately identified
  & \xmark & \cmark & \xmark \\
Whether others' emotions and needs are genuinely understood and responded to
  & \xmark & \cmark & \xmark \\
Whether multi-dimensional information in social contexts is sensitively perceived and accurately responded to
  & \xmark & \cmark & \xmark \\
Whether one can truly read what the other party is thinking
  & \xmark & \cmark & \xmark \\
Whether one's understanding of others' psychology and the essence of relationships is sufficiently deep
  & \xmark & \cmark & \xmark \\
Whether the other party's emotions are genuinely received and steadied within the relationship
  & \xmark & \cmark & \xmark \\
Whether hearts are genuinely soothed and emotional or relational rifts resolved
  & \xmark & \cmark & \xmark \\
Whether the other party's intent and the essence of their question are accurately understood
  & \xmark & \cmark & \xmark \\
Whether the other person is truly kept in mind and thoughtfully looked after
  & \xmark & \cmark & \xmark \\
Whether balance between reason and warmth is maintained and others' needs are responded to sincerely
  & \xmark & \cmark & \xmark \\
Whether others' emotions and circumstances are understood and responded to with authenticity and nuance
  & \xmark & \cmark & \xmark \\
Whether mature and appropriate social competence and the ability to build relationships are demonstrated
  & \xmark & \cmark & \xmark \\
Whether facts are deeply understood and rationally expressed based on evidence
  & \xmark & \cmark & \xmark \\
Whether cognition is continuously recalibrated during social interactions to respond precisely to others' deeper needs
  & \xmark & \cmark & \xmark \\
Whether analyses of others' psychology are both incisive and grounded in reality and proportion
  & \xmark & \cmark & \xmark \\
Whether overt and covert signals in social contexts are sensitively perceived and responded to accurately
  & \xmark & \cmark & \xmark \\
Whether attention is paid to fine details of others' feelings and their latent needs
  & \xmark & \cmark & \xmark \\
Whether others' feelings and circumstances are taken into account during interaction
  & \xmark & \cmark & \xmark \\

\rowcolor{gray!10}
\multicolumn{4}{l}{\footnotesize\textit{\textbf{Behavior Reasoning}}}\\
Whether rational and constructive thinking and decisive action are demonstrated
  & \xmark & \xmark & \cmark \\
Whether win-win outcomes, risks, and long-term goals are balanced under real-world constraints
  & \xmark & \xmark & \cmark \\
Whether things are moved forward to concrete outcomes in a pragmatic and efficient manner
  & \xmark & \xmark & \cmark \\
Whether strategies are flexibly adjusted in dialogue to drive effective and in-depth communication
  & \xmark & \xmark & \cmark \\
Whether principles are upheld while flexibly accommodating all parties' practical realities and relationships
  & \xmark & \xmark & \cmark \\
Whether defensive reactions are triggered, relationships are harmed, and trust is undermined in communication
  & \xmark & \xmark & \cmark \\
Whether effectiveness, proportion, and ethical boundaries are maintained when applying pressure
  & \xmark & \xmark & \cmark \\
Whether actual problems and core concerns are genuinely resolved
  & \xmark & \xmark & \cmark \\
Whether rules, boundaries, and professional integrity commitments are upheld
  & \xmark & \xmark & \cmark \\
Whether de-escalation, ice-breaking, and event progression are advanced in conflict
  & \xmark & \xmark & \cmark \\
Whether collaboration is pursued from a co-building and win-win starting point
  & \xmark & \xmark & \cmark \\
Whether both reducing opposition and driving genuine resolution are balanced in conflict
  & \xmark & \xmark & \cmark \\
Whether goal achievement and relationship maintenance are balanced in collaboration
  & \xmark & \xmark & \cmark \\
Whether logic is clear, expression precise, professionally reliable, thorough, and compelling
  & \xmark & \xmark & \cmark \\
Whether decisions are made boldly, responsibility is accepted, deadlocks are broken, and adaptation occurs in complex situations
  & \xmark & \xmark & \cmark \\
Whether speech is comfortable, natural, direct, and non-hurtful
  & \xmark & \xmark & \cmark \\
Whether the balance among all parties' interests and rule constraints is handled carefully and comprehensively
  & \xmark & \xmark & \cmark \\
Whether various contradictions and conflicts are converted into constructive action
  & \xmark & \xmark & \cmark \\
Whether specific, feasible, principled, and flexible action plans are proposed
  & \xmark & \xmark & \cmark \\
Whether mature, steady, strategic thinking and sincerity are demonstrated in social settings
  & \xmark & \xmark & \cmark \\
Whether appropriate judgment and mature accountability are shown in complex interpersonal situations
  & \xmark & \xmark & \cmark \\
Whether clear strategic awareness and a rational stance are maintained in interpersonal interactions
  & \xmark & \xmark & \cmark \\
Whether sincerity and candor are upheld in expression while also accounting for reality and goodwill
  & \xmark & \xmark & \cmark \\
Whether issues are handled in a professional and rigorous manner
  & \xmark & \xmark & \cmark \\
Whether emotional, rational, and long-term win-win considerations are balanced in conflict
  & \xmark & \xmark & \cmark \\
Whether relationship repair and closure are continuously driven in conflict and collaboration
  & \xmark & \xmark & \cmark \\
Whether strategies and details are flexibly applied to address real-world constraints
  & \xmark & \xmark & \cmark \\
Whether moral proportion and practical effectiveness are balanced in altruistic actions
  & \xmark & \xmark & \cmark \\
Whether various interpersonal relationships are thoughtfully cultivated and maintained
  & \xmark & \xmark & \cmark \\
Whether pressure, manipulation, or concession-based strategies are adopted in interpersonal interactions
  & \xmark & \xmark & \cmark \\
Whether both parties' interests are considered in cooperation and incentive and empowerment tools are flexibly used
  & \xmark & \xmark & \cmark \\
Whether genuine empathy or formulaic probing and scripted packaging are reflected in dialogue
  & \xmark & \xmark & \cmark \\
Whether mature, pragmatic, and far-sighted strategic thinking and action capacity are demonstrated
  & \xmark & \xmark & \cmark \\
Whether the dignity, face, and trust baseline of all parties are respected in interpersonal interactions
  & \xmark & \xmark & \cmark \\
Whether various forms of pressure are applied in response to problems
  & \xmark & \xmark & \cmark \\
Whether problems are genuinely solved in a pragmatic and effective manner
  & \xmark & \xmark & \cmark \\
Whether sincerity, gentleness, and constructiveness are maintained when expressing one's position
  & \xmark & \xmark & \cmark \\
Whether principles are adhered to while practical effectiveness is preserved in responses
  & \xmark & \xmark & \cmark \\
Whether concrete, actionable plans are used to move things forward
  & \xmark & \xmark & \cmark \\
Whether clear boundaries and effective responses are maintained in interpersonal interactions
  & \xmark & \xmark & \cmark \\
Whether the entire interpersonal interaction process is flexibly adapted and optimized
  & \xmark & \xmark & \cmark \\
Whether force is applied precisely and withdrawn steadily in complex interactions
  & \xmark & \xmark & \cmark \\
Whether genuine, pragmatic, and verifiable substantive support is provided sincerely
  & \xmark & \xmark & \cmark \\
Whether dialogues are driven toward practical outcomes with an action-oriented focus
  & \xmark & \xmark & \cmark \\
Whether clear boundaries and authentic proportion are maintained in social interactions
  & \xmark & \xmark & \cmark \\
Whether the other party is guided to participate in co-construction in a respectful and cooperative manner
  & \xmark & \xmark & \cmark \\
Whether honesty and credibility are maintained in factual statements and information construction
  & \xmark & \xmark & \cmark \\
Whether facts are let to speak for themselves and objectivity and truthfulness are upheld
  & \xmark & \xmark & \cmark \\
Whether a pragmatic breakthrough and effective progress can be achieved in crises and deadlocks
  & \xmark & \xmark & \cmark \\
Whether various risks are comprehensively identified, assessed, and appropriately addressed
  & \xmark & \xmark & \cmark \\
Whether a constructive response posture is demonstrated in conflict
  & \xmark & \xmark & \cmark \\
Whether various anxieties and risks are authentically presented and appropriately managed
  & \xmark & \xmark & \cmark \\
Whether solid and flexible logical reasoning ability is demonstrated
  & \xmark & \xmark & \cmark \\
Whether conflict scenarios are flexibly reframed to promote cooperation and mutual gain
  & \xmark & \xmark & \cmark \\
Whether commitments and accountability are expressed clearly and concretely without evasion
  & \xmark & \xmark & \cmark \\
Whether dominance, control, or aggression tendencies are exhibited during interaction
  & \xmark & \xmark & \cmark \\
Whether the establishment, maintenance, and repair of interpersonal relationships are impacted
  & \xmark & \xmark & \cmark \\
Whether multiple strategies are flexibly applied to handle interpersonal conflict and collaborative challenges
  & \xmark & \xmark & \cmark \\
Whether a balance is achieved among rules, needs, values, trust, and experience
  & \xmark & \xmark & \cmark \\
Whether room and openings are reserved for long-term relationships and subsequent development
  & \xmark & \xmark & \cmark \\

\end{longtable}

\end{document}